\documentclass[preprint]{siamonline190516}

\usepackage{lipsum,amsfonts,graphicx,epstopdf,epstopdf,amsopn,cleveref}  
\usepackage{siunitx,tikz}
\usepackage{algorithmic} 
\Crefname{ALC@unique}{Line}{Lines} 
\usepackage[caption=false]{subfig} 
\usepackage{enumitem,lscape}
\usepackage{tablefootnote}
\setlist[enumerate]{leftmargin=.5in}
\setlist[itemize]{leftmargin=.5in}

\definecolor{myblue}{HTML}{1F77b4}
\definecolor{myorange}{HTML}{FF7F0E}
\definecolor{mygreen}{HTML}{2CA02C}
\definecolor{myred}{HTML}{D62728}
\definecolor{mybrown}{HTML}{8C564B}
\definecolor{myyellow}{HTML}{BCBD22}
\definecolor{mycyan}{HTML}{17BECF}

\definecolor{myotherblue}{HTML}{2790F4}
\definecolor{myotherorange}{HTML}{FC7706}

\newcommand{\solidcyan}{\raisebox{2pt}{\protect\tikz{\protect\draw[-,mycyan,solid,line width = 1pt](0,0) -- (16pt,0);}}}
\newcommand{\solidbrown}{\raisebox{2pt}{\protect\tikz{\protect\draw[-,mybrown,solid,line width = 1pt](0,0) -- (16pt,0);}}}
\newcommand{\solidred}{\raisebox{2pt}{\protect\tikz{\protect\draw[-,myred,solid,line width = 1pt](0,0) -- (16pt,0);}}}
\newcommand{\solidyellow}{\raisebox{2pt}{\protect\tikz{\protect\draw[-,myyellow,solid,line width = 1pt](0,0) -- (16pt,0);}}}
\newcommand{\dottedred}{\raisebox{2pt}{\protect\tikz{\protect\draw[-,myred,dash pattern={on 1pt off 1pt},line width = 1pt](0,0) -- (16pt,0);}}}
\newcommand{\dashdottedred}{\raisebox{2pt}{\protect\tikz{\protect\draw[-,myred,dash pattern={on 5pt off 1pt on 1pt off 1pt},line width = 1pt](0,0) -- (16pt,0);}}}
\newcommand{\dashedred}{\raisebox{2pt}{\protect\tikz{\protect\draw[-,myred,dash pattern={on 5pt off 1pt on 5pt off 1pt},line width = 1pt](0,0) -- (16pt,0);}}}

\newcommand{\lhslinedotted}{\raisebox{2pt}{\protect\tikz{\protect\draw[-,mygreen,dash pattern={on 1pt off 1pt},line width = 1pt](0,0) -- (16pt,0);}}}
\newcommand{\lhslinedashed}{\raisebox{2pt}{\protect\tikz{\protect\draw[-,mygreen,dash pattern={on 5pt off 1pt on 5pt off 1pt},line width = 1pt](0,0) -- (16pt,0);}}}
\newcommand{\lhsline}{\raisebox{2pt}{\protect\tikz{\protect\draw[-,mygreen,solid,line width = 1pt](0,0) -- (16pt,0);}}}
\newcommand{\usline}{\raisebox{2pt}{\protect\tikz{\protect\draw[-,myblue,dash pattern={on 5pt off 1pt on 5pt off 1pt},line width = 1pt](0,0) -- (16pt,0);}}}

\newcommand{\uslwline}{\raisebox{2pt}{\protect\tikz{\protect\draw[-,myblue,solid,line width = 1pt](0,0) -- (16pt,0);}}}
\newcommand{\ivrline}{\raisebox{2pt}{\protect\tikz{\protect\draw[-,myorange,dash pattern={on 5pt off 1pt on 5pt off 1pt},line width = 1pt](0,0) -- (16pt,0);}}}
\newcommand{\ivriwline}{\raisebox{2pt}{\protect\tikz{\protect\draw[-,myorange,dash pattern={on 5pt off 1pt on 1pt off 1pt},line width = 1pt](0,0) -- (16pt,0);}}}
\newcommand{\ivrlwline}{\raisebox{2pt}{\protect\tikz{\protect\draw[-,myorange,solid,line width = 1pt](0,0) -- (16pt,0);}}}

\newcommand{\blackline}{\raisebox{2pt}{\protect\tikz{\protect\draw[-,black,solid,line width = 1pt](0,0) -- (16pt,0);}}}
\newcommand{\grayline}{\raisebox{2pt}{\protect\tikz{\protect\draw[-,gray,solid,line width = 1pt](0,0) -- (16pt,0);}}}
\newcommand{\orangeline}{\raisebox{2pt}{\protect\tikz{\protect\draw[-,myotherorange,solid,line width = 1pt](0,0) -- (16pt,0);}}}

\headers{Output-Weighted Optimal Sampling for Experimental Design and UQ}{A. Blanchard and T. Sapsis}

\title{Output-Weighted Optimal Sampling for \\ Bayesian Experimental Design and Uncertainty Quantification\thanks{Submitted to the editors \today.
\funding{This work was supported by the Air Force Office of Scientific Research (MURI Grant No. FA9550-21-1-0058), the Army Research Office (MURI Grant No. W911NF-17-1-0306), and the 2020 MathWorks Faculty Research Innovation Fellowship.
}}}

\author{Antoine Blanchard\thanks{Department of Mechanical Engineering, Massachusetts Institute of Technology, Cambridge, MA 02139 
  (\email{ablancha@mit.edu}, \email{sapsis@mit.edu}).}
\and Themistoklis Sapsis\footnotemark[2]}

\begin{document}

\maketitle

\begin{abstract}
We introduce a class of acquisition functions for sample selection that lead to faster convergence in applications related to Bayesian experimental design and uncertainty quantification.  The approach follows the paradigm of active learning, whereby existing samples of a black-box function are utilized to optimize the next most informative sample. The proposed method aims to take advantage of the fact that  some input directions of the black-box function have a larger impact on the output than others, which is important especially for systems exhibiting rare and extreme events. The acquisition functions introduced in this work leverage the properties of the likelihood ratio, a quantity that acts as a probabilistic sampling weight and guides the active-learning algorithm towards regions of the input space that are deemed most relevant.  We demonstrate the proposed approach in the uncertainty quantification of a hydrological system as well as the probabilistic quantification of rare events in dynamical systems and the identification of their precursors in up to 30 dimensions. 
\end{abstract}



\section{Introduction}
\label{sec:1}

Modern societies have reached such high levels of sophistication that real-world systems (e.g., social networks, financial markets, biological systems, artificial-intelligence algorithms) have become far too intricate to design, optimize and analyze using traditional techniques.  Conceptually, these systems can be viewed as input--output relationships, with the output representing some quantity of interest (e.g., drag force, temperature, wave height, or stock price) and the inputs carrying with them some degree of uncertainty related to experimental or computational parameters.  The input--output relationship, therefore, is a ``black box'' which can be learned using standard machine-learning techniques by making a series of queries and fitting a statistical model to the resulting input--output pairs.

In many practical applications, each query may take days or even weeks to produce a result (e.g., the black box is a massively parallel computer code or an on-site experiment).  Therefore, each query point must be selected gingerly, otherwise time and resources will be wasted.  This difficulty is exacerbated when the black box has a large number of input parameters (i.e., it is high-dimensional) and possesses strongly nonlinear features.  If, in addition, the black box has the ability to generate extreme events (i.e., events that combine high-magnitude impact with low frequency of occurrence \cite{albeverio2006extreme,bayarri2009using,lucarini2016extremes}), then quantifying the statistics of those extremes through economical sampling of the input space is quite daunting.

The question of sample selection in uncertainty quantification of black-box functions
is an active area of research.  In randomized-sampling methods, samples are drawn from a distribution that has been optimized in some way.  For example, importance sampling utilizes a biasing distribution to concentrate sampling on the regions of the input space that generate extreme outcomes \cite{kahn1953methods,shinozuka1983basic}.  In practice, constructing the optimal biasing distribution is quite challenging, and approximations based on large-deviation theory \cite{dematteis2019extreme}, the cross-entropy method \cite{uribe2020cross}, or other techniques \cite{wahal2019bimc} are often inevitable.  Another example is the subset-simulation approach \cite{au2001estimation}, where the probability of a rare event is expressed as a product of larger conditional  probabilities computed by Markov chain Monte Carlo simulation.

On the other side of the spectrum we have active-learning methods which aim to optimize the selection of each individual sample instead of drawing them in bulk from a carefully crafted distribution \cite{sacks1989design,chaloner1995bayesian,gramacy2009adaptive,choe2018uncertainty}.  Few studies have addressed the question of active learning for rare-event quantification, and those who did focused primarily on the estimation of a particular quantile of the output distribution and therefore could not characterize the full distribution \cite{oakley2004estimating,picard2013rare,jala2014sequential,schobi2017rare}.  Recently, Mohamad and Sapsis \cite{mohamad2018sequential} proposed an algorithm that accelerates convergence of the output statistics compared to  approaches based on uncertainty sampling or mutual information.  However, the optimization process involved in their approach is computationally expensive, limiting applicability to low-dimensional input spaces.

A critical issue in active learning is the choice of acquisition function, i.e., the criterion used to select which sample to query next in an optimal manner.  In goal-oriented uncertainty quantification, the choice of acquisition function largely depends on the features of the quantity of interest one wishes to identify (e.g., the statistical expectation of the output \cite{pandita2019bayesian}, the tails of the output distribution \cite{mohamad2018sequential}, or the maximum value that the black-box function can produce \cite{stefanakis2014can}).  Acquisition functions come in various shapes and forms \cite{chaloner1995bayesian,shahriari2015taking}, but many popular criteria suffer from severe limitations, including high computational cost, intractability in high dimensions, and inability to discriminate between active and idle input variables \cite{sapsis2020output}.

We introduce a class of acquisition functions for active learning of black-box functions specifically designed for systems capable of generating extreme events.  As in \cite{mohamad2018sequential} and \cite{sapsis2020output}, the proposed criteria guide the search algorithm toward regions of the input space that are associated with unusual output values associated with rare events.  Unlike \cite{mohamad2018sequential}, their computational complexity is comparable with that of traditional active-learning approaches, and they can be approximated in a way that makes the approach tractable in high dimensions.  Unlike \cite{sapsis2020output}, they are derived for a much larger class of models (i.e., nonlinear, nonparametric models).  In addition, the proposed criteria contain a mechanism that accounts for the importance of the output relative to the input, and therefore are not limited to rare-event quantification but can be applied to any problem related to experimental design and uncertainty quantification.

The remainder of the paper is structured as follows.  We formulate the problem in \cref{sec:2}, present the sampling approach in \cref{sec:3}, followed by numerical results in \cref{sec:4} and conclusions in \cref{sec:5}.

\section{Formulation of the Problem}
\label{sec:2}

\subsection{Active Learning of Black-Box Functions}
\label{sec:21}

We consider a function $f : \mathbb{R}^d \longrightarrow \mathbb{R}$ defined over a compact set $\mathcal{X} \subset \mathbb{R}^d$.  The function $f$ is treated as a black box, having the following properties: a) it has no simple closed form, and neither do its gradients; b) its internal structure (e.g., linearity or concavity) is unknown; and c) it can be queried at any arbitrary point $\mathbf{x} \in \mathcal{X}$, each evaluation producing a potentially noise-corrupted output
\begin{equation}
y = f(\mathbf{x}) + \varepsilon, \quad \varepsilon \sim \mathcal{N}(0,  \sigma_\varepsilon^2),
\label{eq:21}
\end{equation}
where uncertainty in observations is modeled with additive Gaussian noise.  To avoid pathological cases we require that $f$ be Lipschitz continuous \cite{brochu2010tutorial}.  

In practice it is difficult to uncover the internal workings of the black-box function because each query can be very costly, either being time-consuming or requiring vast amounts of resources.  Such is the case when $f$ is a machine-learning algorithm (with $\mathbf{x}$ the hyper-parameters), a large-scale computer simulation of a physical system (with $\mathbf{x}$ the physical parameters), or a field experiment (with $\mathbf{x}$ the experimental parameters).  It is thus clear that a brute-force approach, in which the objective function $f$ would be queried at a large number of input points, is not possible.   

To combat this, one approach is to proceed sequentially.  Starting from an initial dataset of input--output pairs, Algorithm \ref{alg:1} iteratively probes the input space $\mathcal{X}$ and, with each point visited, attempts to construct a surrogate model (or ``emulator'') for the objective function.  At each iteration the ``best next point'' to visit is selected meticulously by minimizing an acquisition function $a : \mathbb{R}^d \longrightarrow \mathbb{R}$ which serves as a beacon for the algorithm as it scouts the input space.  After a specified number of iterations, the algorithm returns the surrogate model it has constructed, which can then be used in analyses as a substitute for the unknown black-box function $f$.

\begin{algorithm}[tb]
   \caption{Sequential search for active learning of black-box functions.}
   \label{alg:1}
\begin{algorithmic}[1]
   \STATE {\bfseries Input:} Number of iterations $n_\textit{iter}$ 
   \STATE {\bfseries Initialize:} Surrogate model $\bar{f}$ on initial dataset of input--output pairs $\mathcal{D}_0 = \{\mathbf{x}_i, y_i\}_{i=1}^{n_\textit{init}}$
   \FOR{$n=1$ {\bfseries to} $n_\textit{iter}$}
      \STATE Select best next point $\mathbf{x}_{n}$ by minimizing acquisition function $a$:
      	\begin{equation*}
	\mathbf{x}_{n} = \operatorname*{arg\,min}_{\mathbf{x} \in \mathcal{X}} a(\mathbf{x}; \bar{f}, \mathcal{D}_{n-1})
	\end{equation*}
	\STATE Evaluate objective function $f$ at $\mathbf{x}_{n}$ and record $y_{n}$
	\STATE Augment dataset: $\mathcal{D}_{n} = \mathcal{D}_{n-1} \cup \{\mathbf{x}_{n} , y_{n} \}$
   \STATE Update surrogate model
   \ENDFOR
   \RETURN Final surrogate model
\end{algorithmic}
\end{algorithm}

Algorithm \ref{alg:1} is at the foundation of Bayesian experimental design (BED) \cite{chaloner1995bayesian} and Bayesian optimization (BO) \cite{shahriari2015taking}.  In BED, the goal is to learn the objective function \textit{globally} in order to make accurate predictions in locations where $f$ has not been observed, or to compute a global quantity of interest such as the pdf of the output or the integral of the objective function over the input space.  In BO, the focus is on finding the global \textit{minimum} of the objective function, with no consideration for the behavior of $f$ away from the minimizer.  Both BED and BO problems can be addressed with the sequential approach in Algorithm \ref{alg:1}, with the only difference between BED and BO being the choice of acquisition function.

The acquisition function is a crucial component of the sequential algorithm as it guides  exploration of the input space and prescribes the points at which the objective function is queried.  Therefore, the acquisition function should reflect the type of problem one is attempting to solve.  It should favor points that most improve the quality of the surrogate model globally for BED, and in the immediate vicinity of the global minimizer for BO.  The final recommendation likewise depends on the problem at hand.  In BED it is the surrogate model itself, whereas in BO it is the minimizer of the surrogate model
\begin{equation}
\mathbf{x}^* = \operatorname*{arg\,min}_{\mathbf{x} \in \mathcal{X}} \bar{f}(\mathbf{x}).
\label{eq:22}
\end{equation}

That Algorithm \ref{alg:1} can be used to solve two fundamentally different problems---BED and BO---is important for two reasons.  First, it simplifies implementation, allowing use of the same computer code in either situation provided the appropriate acquisition functions are available.  Second, it highlights the fact that the main difference between BED and BO lies in the choice of acquisition function, with the key issue being the trade-off between exploration and exploitation.  Acquisition functions used in BED should encourage the algorithm to visit regions of the input space where uncertainty is high (i.e., pure exploration), thus improving the quality of the surrogate model \textit{globally}.  By contrast, acquisition functions used in BO should seek a compromise between exploration and exploitation, with the latter promoting regions where the surrogate model predicts small values.  

The connection between BED and BO suggests the possibility of ``repurposing'' a purely explorative BED acquisition function into one which is suitable for BO, i.e., more aggressive towards minima \cite{srinivas2009gaussian}.  Since the focus of this work is on experimental design, this idea will not be pursued any further.  But this issue is worth mentioning because it suggests that any successful BED strategy introduced in this paper has the potential of being equally successful in the context of BO \cite{blanchard2020bayesian}.

\subsection{Model Selection}
\label{sec:22}

The use of the adjective ``Bayesian'' in \cref{sec:21} has to do with the other key issue in Algorithm \ref{alg:1}, namely, model selection.  The choice of surrogate model is important because it encapsulates our belief about what the objective function looks like given the data collected by the algorithm.  Bayesian approaches have the advantage that they allow a rigorous treatment of uncertainty in observations, with the model being described in terms of probability distributions.  Use of prior and posterior distributions makes it possible to continually update the surrogate model as more information becomes available.

In this work we use a non-parametric Bayesian approach based on Gaussian process (GP) regression \cite{rasmussen2006gaussian}.  This is appropriate because GPs a) are agnostic to the details of the black box, b) provide a way to quantify uncertainty associated with noisy observations, and c) are robust, versatile, easy to implement, and relatively inexpensive to train.  A Gaussian process $\bar{f}(\mathbf{x})$ is completely specified by its mean function $m(\mathbf{x})$ and covariance function $k(\mathbf{x},\mathbf{x}')$.  For a dataset $\mathcal{D}$ of input--output pairs (written in matrix form as $\{\mathbf{X}, \mathbf{y}\}$) and a Gaussian process with constant mean $m_0$, the random process $\bar{f}(\mathbf{x})$ conditioned on $\mathcal{D}$ follows a normal distribution with posterior mean and variance
\begin{subequations}
\begin{gather}
\mu(\mathbf{x}) = m_0 + k(\mathbf{x}, \mathbf{X}) \mathbf{K}^{-1} (\mathbf{y} -m_0), \label{eq:23a}\\
\sigma^2(\mathbf{x}) = k(\mathbf{x},\mathbf{x}) - k(\mathbf{x}, \mathbf{X}) \mathbf{K}^{-1} k(\mathbf{X},\mathbf{x}),\label{eq:23b}
\end{gather}
\end{subequations}
respectively, where $\mathbf{K} = k(\mathbf{X},\mathbf{X})+ \sigma_\varepsilon^2 \mathbf{I}$.  \Cref{eq:23a} can be used to predict the value of the surrogate model at any point $\mathbf{x}$, and \cref{eq:23b} to quantify uncertainty in prediction at that point \cite{rasmussen2006gaussian}. 

In GP regression, the choice of covariance function is important, and in this work we use the radial-basis-function (RBF) kernel with automatic relevance determination (ARD),
\begin{equation}
k(\mathbf{x},\mathbf{x}') = \sigma_f^2 \exp [ -(\mathbf{x} - \mathbf{x}')^\mathsf{T} \mathbf{\Theta}^{-1}(\mathbf{x} - \mathbf{x}') /2],
\label{eq:24}
\end{equation}
where $\mathbf{\Theta}$ is a diagonal matrix containing the lengthscales for each dimension.  The advantage of the RBF kernel will become clear in the next sections.  For a given dataset, the GP hyper-parameters appearing in the covariance function ($\sigma_f^2$ and $\mathbf{\Theta}$ in \cref{eq:24}) are trained by maximum likelihood estimation.  

Exact inference in GP regression requires inverting the matrix $\mathbf{K}$, typically at each iteration.  This is usually done by Cholesky decomposition whose cost scales like $O(n^3)$, with $n$ the number of observations \cite{rasmussen2006gaussian,shahriari2015taking}.  (A cost of $O(n^2)$ can be achieved if the parameters of the covariance function are kept fixed.)  Although an $O(n^3)$ scaling may seem formidable, it is important to note that in BED (and BO for that matter) the number of observations rarely exceeds a few dozens (or perhaps a few hundreds), as an unreasonably large number of observations would defeat the whole purpose of the sequential algorithm.

\subsection{Acquisition Functions for Bayesian Experimental Design}
\label{sec:23}

As discussed in \cref{sec:21}, the acquisition function plays a critical role in Algorithm \ref{alg:1} as it is the sole decider of where to query the black-box function.  In BED, the role of the acquisition function is to reduce uncertainty \textit{globally} in that the surrogate model should approximate the black-box function reasonably well across the whole input space.  This can be achieved in a number of ways, which we review below in order of increasing complexity.  

\subsubsection*{Uncertainty sampling} The most intuitive approach is to select the best next point where the predictive variance of the GP model is the highest:
\begin{equation}
a_\textit{US}(\mathbf{x}) = \sigma^2(\mathbf{x}).
\label{eq:25}
\end{equation}
Uncertainty sampling (US) ensures that model uncertainty is distributed somewhat evenly over the input space.  The popularity of US can be largely explained by the facts that a) implementation is straightforward, b) evaluation is inexpensive, and c) gradients are analytic, making possible the use of gradient-based optimizers.  The combination of these three features makes the search for the best next point considerably more efficient than otherwise.  This is important because a key prerequisite for \cref{alg:1} to provide any sort of advantage over a brute-force approach is that the cost of optimizing the acquisition function should be small compared to that of querying the black-box function $f$.

In the BED literature, US is also known as the active-learning-MacKay (ALC) algorithm \cite{gramacy2009adaptive}.  As discussed by MacKay \cite{mackay1992information}, selecting the point that maximizes the predictive variance is approximately equivalent to maximizing the information gained about model parameters upon addition of that point.  It has also been reported that US has a disproportionate tendency toward selecting points on the boundary of the search space, simply because the variance is often largest far away from regions where data has been collected.  Whether or not boundary points are informative is still an open question \cite{krause2008near,siivola2018correcting}.

\subsubsection*{Integrated variance reduction}  Instead of considering only past observations, one may investigate the effect of observing a hypothetical ``ghost'' point $\mathbf{x}$ on the overall model variance \cite{cohn1994neural}.  This effect is measured by 
\begin{equation}
a_\textit{IVR}(\mathbf{x}) = \int \left[ \sigma^2(\mathbf{x}') - \sigma^2(\mathbf{x}'; \mathbf{x}) \right] \mathrm{d}\mathbf{x}',
\label{eq:26}
\end{equation}
where $\sigma^2(\mathbf{x}'; \mathbf{x})$ is the predictive variance at $\mathbf{x}'$ had the ghost point $\mathbf{x}$ been observed with output $\mu(\mathbf{x})$.  Therefore, maximizing IVR has the effect of maximally reducing the overall model variance.  We note that others have referred to IVR as the active-learning-Cohn (ALC) algorithm \cite{gramacy2009adaptive}.  

However, \cref{eq:26} suffers from several shortcomings.  First, it involves an integral over the input space.  Second, computation of $\sigma^2(\mathbf{x}'; \mathbf{x})$ requires updating the covariance matrix $\mathbf{K}$ for every ghost point considered, with expectations about the associated computational cost being similar to those described in \cref{sec:22}.  The combination of these two factors makes evaluation of IVR cumbersome, and that of its gradients even more so.  To eliminate these issues, we first note that 
\begin{equation}
a_\textit{IVR}(\mathbf{x}) = \frac{1}{\sigma^2(\mathbf{x})} \int \mathrm{cov}^2(\mathbf{x}, \mathbf{x}') \, \mathrm{d}\mathbf{x}',
\label{eq:27}
\end{equation}
where $\mathrm{cov}(\mathbf{x}, \mathbf{x}') = k(\mathbf{x},\mathbf{x}') -k(\mathbf{x},\mathbf{X}) \mathbf{K}^{-1} k(\mathbf{X},\mathbf{x}')$ is the posterior covariance between $\mathbf{x}$ and $\mathbf{x}'$  (\cref{app:1}).  Now, evaluation of $a_\textit{IVR}$ merely requires a mechanism to compute 
\begin{equation}
\hat{k}(\mathbf{x}_1, \mathbf{x}_2) = \int k(\mathbf{x}_1, \mathbf{x}) k( \mathbf{x}, \mathbf{x}_2)\, \mathrm{d}\mathbf{x}
\label{eq:28}
\end{equation}
for any $\mathbf{x}_1$ and $\mathbf{x}_2$.  For the RBF kernel, \cref{eq:28} and its gradients can be computed analytically, and consequently the same is true for IVR and its gradients (\cref{app:2}). 

\subsubsection*{Mean model error}  In BED it is common for the input space to be equipped with a non-uniform probability density function $p_\mathbf{x}$ which reflects uncertainty in the input.  (Oftentimes the prior $p_\mathbf{x}$ is a given.  When it is not, it can be retrieved from data by approximating the joint density of the input variables with a kernel density estimator.)  It is therefore natural to bias the search towards regions of the input space that are more likely and realistic than others \cite{sacks1989design}.  This is done by incorporating $p_\mathbf{x}(\mathbf{x})$ as a sampling weight in \cref{eq:27}, leading to
\begin{equation}
a_\textit{IVR-IW}(\mathbf{x}) = \frac{1}{\sigma^2(\mathbf{x})} \int \mathrm{cov}^2(\mathbf{x}, \mathbf{x}') p_\mathbf{x}(\mathbf{x}') \, \mathrm{d}\mathbf{x}',
\label{eq:29}
\end{equation}
where the suffix ``IW'' stands for ``input-weighted''.  Specifically, IVR-IW measures the expected variance reduction resulting from the addition of ghost point $\mathbf{x}$ to the dataset, since by virtue of the equality between \cref{eq:26} and \cref{eq:27} it holds that
\begin{equation}
a_\textit{IVR-IW}(\mathbf{x}) = \mathbb{E}_{\mathbf{x}'}[ \sigma^2(\mathbf{x}') - \sigma^2(\mathbf{x}'; \mathbf{x}) ].
\label{eq:210}
\end{equation}

From a computational standpoint, the introduction of the sampling weight $p_\mathbf{x}(\mathbf{x})$ makes evaluation of IVR-IW more challenging than that of IVR because now we need a mechanism to compute 
\begin{equation}
\hat{k}_p(\mathbf{x}_1, \mathbf{x}_2) = \int k(\mathbf{x}_1, \mathbf{x}) k( \mathbf{x}, \mathbf{x}_2) p_\mathbf{x}(\mathbf{x}) \, \mathrm{d}\mathbf{x}.
\label{eq:211}
\end{equation}
The above integral and its gradients are analytic only in specific cases; for example, when $p_\mathbf{x}$ is Gaussian and $k(\mathbf{x},\mathbf{x}')$ is the RBF kernel.  For more complicated cases, IVR-IW should be evaluated by Monte Carlo integration, limiting applicability to low-dimensional problems and prohibiting use of gradient-based optimization routines.  

\subsubsection*{Mutual information}  Formally, the expected information gain resulting from a new data point being observed is quantified by the expected Kullback--Leibler (KL) divergence between the prior and posterior distributions:
\begin{equation}
a_\textit{MI}(\mathbf{x}) = \iint \log \frac{p_\mathbf{x}(\mathbf{x}' | y, \mathbf{x})}{p_\mathbf{x}(\mathbf{x}')} p_y(y, \mathbf{x}' | \mathbf{x})\,\mathrm{d}\mathbf{x}'\, \mathrm{d} y.
\label{eq:212}
\end{equation}
Maximizing the expected KL divergence is equivalent to maximizing the entropy transfer or mutual information (MI) between input and output when the new point $\mathbf{x}$ is received \cite{chaloner1995bayesian}.  The major drawback of MI is that it has no closed form, unless the surrogate model and prior distribution cooperate (e.g., linear regression with Gaussian prior, in which case MI reduces to US \cite{chaloner1995bayesian}).  As a result, MI must be approximated by Monte Carlo integration, with limitations on applicability being similar to those discussed earlier for IVR-IW.\\

While other approaches exist, either they are computationally more complex or they have narrower applicability than those previously discussed.  For example, the expected improvement for global fit (EIGF) augments US with a term favoring points for which the predicted output is furthest away from previously seen values \cite{lam2008sequential}; but EIGF outperforms US only in very specific cases \cite{maljovec2013adaptive}.  Likewise, several variations of MI have been proposed \cite{beck2016sequential,wang2017max}, but they often entail heavy sampling requirements, intractability in high dimensions, or limited choice of GP kernels.  

Another key issue is that traditional acquisition functions (including those discussed above) fail to take into account information about the output space available from previous observations.  (For example, note how \cref{eq:25,eq:26,eq:29} lack a dependence on $y$ or $\mu$, resulting in the algorithm being unable to discriminate directions that have no effect on the output from those that do.)  When the output space \textit{is} taken into account, it is at the expense of computational efficiency, with costly entropy estimations often being inevitable \cite{hoffman2015output,beck2016sequential}.  This is important because previous studies suggest that incorporating information about the output space can lead to significant gains when the objective function is noisy, multi-modal, and has the ability to generate rare events.

In light of this, our goal is to design acquisition functions that put a premium on the output values of previously visited data points while being computationally tractable.  This is explored in the next section.

\section{Bayesian Experimental Design with Output-Weighted Optimal Sampling}
\label{sec:3}

\subsection{The Significance of the Likelihood Ratio}
\label{sec:31}

To quantify the importance of the output relative to the input, we draw inspiration from the theory of importance sampling and introduce the
\textit{likelihood ratio}
\begin{equation}
w(\mathbf{x}) = \frac{p_\mathbf{x}(\mathbf{x})}{p_\mu(\mu(\mathbf{x}))},
\label{eq:31}
\end{equation} 
where $p_\mu$ denotes the pdf of the GP posterior mean conditioned on the input.  In the importance-sampling literature, $p_\mathbf{x}$ is referred to as the ``nominal distribution'', and $p_\mu$ as the ``importance distribution''.  The meaning of the latter might be difficult to grasp considering that the GP mean $\mu$ is a \textit{deterministic} function of $\mathbf{x}$.  The key is to view $\mathbf{x}$ as a random variable, distributed according to the prior $p_\mathbf{x}$.  The GP mean $\mu$ is, in turn, a random variable, and we let $p_\mu$ denote its density.  (We will expand on the computation of $p_\mu(\mu(\mathbf{x}))$ in \cref{sec:33}.)

The likelihood ratio is important in cases where some points are more important than others in determining the value of the output.  It acts as a \textit{sampling weight}, assigning to each data point $\mathbf{x} \in \mathcal{X}$ a measure of ``relevance'' defined in probabilistic terms.  For points with similar probability of being observed ``in the wild'' (i.e., same $p_\mathbf{x}$), the likelihood ratio assigns more weight to those that have a large impact on the magnitude of output (i.e., small $p_\mu$).  For points with similar impact on the output (i.e., same $p_\mu$), it promotes those with higher probability of occurrence (i.e., large $p_\mathbf{x}$).  In other words, it favors points for which the magnitude of the output is \textit{unusually large} over points associated with frequent, average output values.

\subsection{Likelihood-weighted acquisition functions}
\label{sec:32}

When the likelihood ratio appears in an acquisition function, we refer to the latter as a ``likelihood-weighted'' (LW) acquisition function.  To the best of our knowledge, the only LW acquisition function that has been proposed for BED is the so-called Q criterion \cite{sapsis2020output}:
\begin{equation}
a_\mathit{Q}(\mathbf{x}) = \int \sigma^2(\mathbf{x}'; \mathbf{x}) w(\mathbf{x}') \, \mathrm{d}\mathbf{x}'.
\label{eq:32}
\end{equation}
Heuristically, the Q criterion can be thought of as a variant of IVR-IW in which the likelihood ratio is substituted for the input pdf, allowing the algorithm to give more weight to unusual output values.  The derivation in \cref{app:1} can indeed be adapted to prove that minimizing $a_\mathit{Q}$ is strictly equivalent to maximizing
\begin{equation}
a_\textit{IVR-LW}(\mathbf{x}) = \frac{1}{\sigma^2(\mathbf{x})} \int \mathrm{cov}^2(\mathbf{x}, \mathbf{x}') w(\mathbf{x}') \, \mathrm{d}\mathbf{x}'.
\label{eq:33}
\end{equation}

This interpretation is intuitive but lacks rigor because the substitution of $w(\mathbf{x})$ for $p_\mathbf{x}(\mathbf{x})$ might seem arbitrary.  To establish the credibility of the Q criterion, we must first recognize that it has ramifications running deep within the field of rare-event quantification.  This connection was found by Sapsis \cite{sapsis2020output} who proved that the Q criterion is an upper bound for 
\begin{equation}
a_L(\mathbf{x}) = \int \left| \log p_{\mu_+}(y) - \log p_{\mu_-}(y) \right| \mathrm{d}y,
\label{eq:34}
\end{equation}
where $\mu_{\pm}(\mathbf{x}'; \mathbf{x}) =  \mu(\mathbf{x}') \pm \sigma^2(\mathbf{x}'; \mathbf{x})$.  The criterion $a_L$ was designed specifically for rare-event quantification, as evidenced by the use of logarithms for the output pdf \cite{mohamad2018sequential}.  With \cref{eq:34} the best next point is selected so as to most reduce uncertainty in the output pdf, with extra emphasis on the tails.

Preferring \cref{eq:33} to \cref{eq:34} has two advantages.  First, evaluation of \cref{eq:34} is quite slow and tedious, and its gradients cannot be computed in closed form; in contrast, we will see in \cref{sec:33} that \cref{eq:33} and its gradients can be computed very efficiently, even in high dimensions.  Second, the unexpected connection between the Q criterion and the IVR-type acquisition functions implies that for all $\mathbf{x}\in \mathcal{X}$,
\begin{equation}
0 \leq a_\textit{IVR-LW}(\mathbf{x}) \leq M a_{IVR}(\mathbf{x}),
\label{eq:350}
\end{equation}
where $M$ is a positive constant that bounds $w(\mathbf{x})$ from above.  In the limit of many observations, $a_{IVR}(\mathbf{x})$ goes to zero, which establishes convergence of IVR-LW.

Since the likelihood ratio may be viewed as a sampling weight, we also introduce the likelihood-weighted counterpart to US,
\begin{equation}
a_\textit{US-LW}(\mathbf{x}) = \sigma^2(\mathbf{x}) w(\mathbf{x}),
\label{eq:35}
\end{equation}
which directs the algorithm towards uncertain yet highly ``relevant'' regions of the input space.  Nothing is known about the behavior and properties of US-LW as this acquisition function has not been introduced previously.  In the remainder of the paper, we will limit our investigation of LW acquisition functions to US-LW and the Q criterion.

\subsection{Computation of LW acquisition functions}
\label{sec:33}

We must ensure that the introduction of the likelihood ratio does not compromise our ability to compute the acquisition functions efficiently.  The two key issues are the computation of the likelihood ratio itself, and the evaluation of the integral appearing in \cref{eq:33}.  It is also important that the gradients of the acquisition functions be tractable to allow use of gradient-based optimizers.

We first note that to evaluate the likelihood ratio, we must estimate the conditional pdf of the posterior mean $p_\mu$, typically at each iteration.  This can be done by computing $\mu(\mathbf{x})$ for a large number of input points and applying KDE to the resulting samples (\cref{alg:3}).  Fortunately, KDE is to be performed in the (one-dimensional) output space, allowing use of fast FFT-based algorithms which scale linearly with the number of samples \cite{fan1994fast}.  We also note that the gradients of $w(\mathbf{x})$ are given by
\begin{equation}
\frac{\mathrm{d} w(\mathbf{x})}{\mathrm{d} \mathbf{x}} = \frac{1}{p_\mu(\mu(\mathbf{x}))^2} \left[ \frac{\mathrm{d} p_\mathbf{x}(\mathbf{x})}{\mathrm{d} \mathbf{x}} p_\mu(\mu(\mathbf{x})) - \frac{\mathrm{d} p_\mu(y)}{\mathrm{d}y} \frac{\mathrm{d} \mu(\mathbf{x})}{\mathrm{d} \mathbf{x}} p_\mathbf{x}(\mathbf{x}) \right]\!,
\label{eq:36}
\end{equation}
where $\mathrm{d} p_\mu(y)/\mathrm{d} y$ can be approximated efficiently by finite differences, and $\mathrm{d} \mu(\mathbf{x})/\mathrm{d} \mathbf{x}$ can be computed analytically using the GP expression \cref{eq:23a} \cite{mchutchon2013differentiating}.  With this in hand, US-LW and its gradients can be computed analytically.

\begin{algorithm}[tb]
   \caption{Computation of the importance distribution}
   \label{alg:3}
\begin{algorithmic}[1]
   \STATE {\bfseries Input:} Posterior mean $\mu$, input prior $p_\mathbf{x}$
   
\algorithmicdo\ 
\begin{ALC@g}
     \STATE Draw a large number of samples $\{\mathbf{x}_k\}_{k=1}^M$, either uniformly or from the prior $p_\mathbf{x}$
     \STATE Push the samples through the GP model and obtain $\{\mu_k\}_{k=1}^M$, with $\mu_k = \mu(\mathbf{x}_k)$
     \STATE  Perform kernel density estimation on the collection $\{\mu_k\}_{k=1}^M$ to obtain the (univariate) output density $p_\mu(y)$, $y$ a dummy variable
     \STATE  For a particular $\mathbf{x}$, back out the value of $p_\mu(\mu(\mathbf{x}))$ by looking up the value of $p_\mu(y)$ corresponding to $y=\mu(\mathbf{x})$
     
\end{ALC@g}
\algorithmicend~\algorithmicdo\
   \RETURN Importance distribution $p_\mu(\mu(\mathbf{x}))$
\end{algorithmic}
\end{algorithm}

For the Q criterion, approximating the likelihood ratio is inevitable if one wants to make the integral in \cref{eq:33} analytic.  For example, Sapsis \cite{sapsis2020output} used a quadratic approximation for $1/p_\mu$, a trick that allowed analytical computation of the Q criterion and its gradients but also led to drastic restrictions on the form of $p_\mathbf{x}$ and $p_\mu$ and on the type of surrogate model.  To eliminate these requirements, we approximate $w(\mathbf{x})$ with a Gaussian mixture model (GMM):
\begin{equation}
w(\mathbf{x})  \approx \sum_{i=1}^{n_\mathit{GMM}} \alpha_i \, \mathcal{N}(\mathbf{x}; \boldsymbol{\omega}_i, \mathbf{\Sigma}_i).
\label{eq:37}
\end{equation}

The GMM approximation has two advantages.  First, when combined with the RBF kernel, the problematic integral in \cref{eq:33} and its gradients become analytic (\cref{app:3}).  Second, the GMM approximation imposes no restriction on the nature of $p_\mathbf{x}$ and $p_\mu$, unlike in \cite{sapsis2020output}.  The number of Gaussian mixtures to be used in \cref{eq:37} is at the discretion of the user.  It may be kept constant throughout the search or modified ``on the fly'', either according to a pre-defined schedule or by selecting the value of $n_\mathit{GMM}$ that minimizes the Akaike information criterion (AIC) or the Bayesian information criterion (BIC) at each iteration or less frequently if so desired \cite{vanderplas2016python}.

\begin{figure}[tbhp]
\centering 
\subfloat[Oakley--O'Hagan function]{\label{fig:1a}\includegraphics[width=6in]{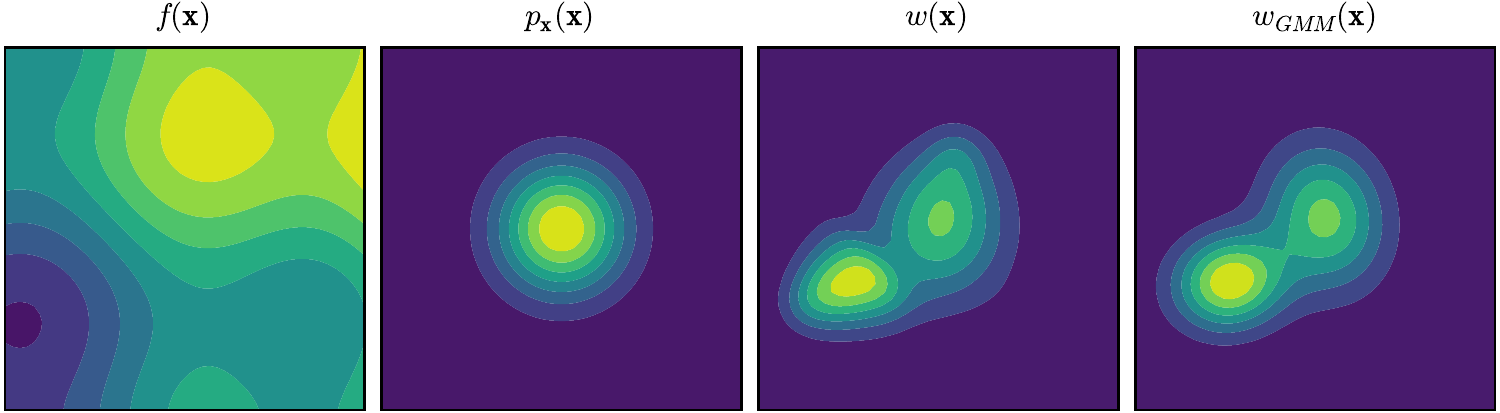}} 

\subfloat[Michalewicz function]{\label{fig:1b}\includegraphics[width=6in]{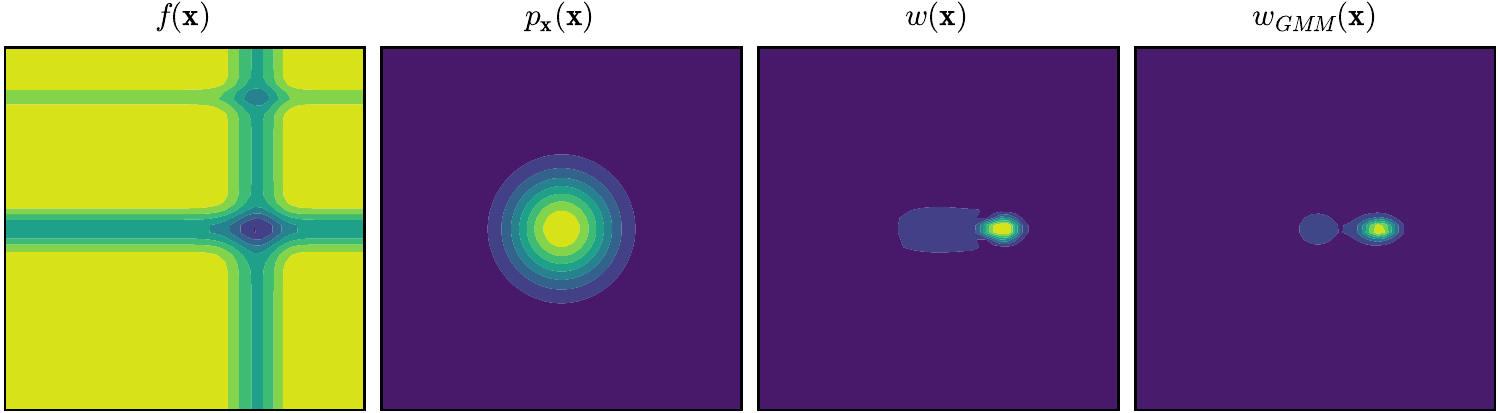}} 
\caption{From left to right: Contour plots of the objective function $f(\mathbf{x})$, the input pdf $p_\mathbf{x}(\mathbf{x})$, the likelihood ratio $w(\mathbf{x})$ with $f$ in place of $\mu$ in \cref{eq:31}, and the GMM approximation of the likelihood ratio $w_\mathit{GMM}(\mathbf{x})$ with two Gaussian mixtures.}
\label{fig:1}
\end{figure}

For a simple illustration of the benefits provided by the likelihood ratio, we consider the Oakley--O'Hagan function \cite{oakley2002bayesian}
\begin{subequations}
\begin{equation}
f(\mathbf{x}) = 5 + x_1 + x_2 + 2 \cos(x_1) + 2\sin(x_2)
\label{eq:38a}
\end{equation}
with $\mathbf{x} \in [-4,4]^2$ and $p_\mathbf{x}(\mathbf{x}) = \mathcal{N}(\mathbf{0}, \mathbf{I})$, and the 2-D Michalewicz function
\begin{equation}
f(\mathbf{x}) = - \sin(x_1) \sin^{20}(x_1^2/\pi) - \sin(x_2) \sin^{20}(2x_2^2/\pi)
\label{eq:38b}
\end{equation}%
\label{eq:38}%
\end{subequations}%
with $\mathbf{x} \in [0,\pi]^2$ and $p_\mathbf{x}(\mathbf{x}) = \mathcal{N}(\mathbf{0}+\pi/2, 0.1\mathbf{I})$.  For these functions, \cref{fig:1} makes it visually clear that the likelihood ratio gives more emphasis to the regions where the magnitude of the objective function is unusually large and promotes points that have a large impact on the output despite their probability of occurrence not being the largest. \Cref{fig:1} also shows that $w(\mathbf{x})$ can be approximated satisfactorily with a small number of Gaussian mixtures, a key prerequisite for preserving algorithm efficiency.

We summarize our algorithm for computation of likelihood-weighted acquisition functions in \cref{alg:2}.  We note that by virtue of \cref{eq:36}, the GMM step in \cref{alg:2} is optional for US-LW.  However, it is useful to compare the performance of US-LW with and without the GMM approximation in order to assess the extent to which the GMM approximation captures the important features of $w(\mathbf{x})$.  We also note that up to now we have adhered to conventional notation for the acquisition functions at the expense of making it clear whether they should be minimized or maximized; \cref{tab:1} should dissipate any ambiguity.

\begin{algorithm}[tb]
   \caption{Likelihood-weighted acquisition function}
   \label{alg:2}
\begin{algorithmic}[1]
   \STATE {\bfseries Input:} Posterior mean $\mu$, input prior $p_\mathbf{x}$, functional form for LW acquisition function $a(\mathbf{x}; w(\mathbf{x}))$, number of Gaussian mixtures $n_\mathit{GMM}$

\algorithmicdo\ 
\begin{ALC@g}
     \STATE Compute the importance distribution $p_\mu$ using \cref{alg:3}
     \STATE Compute  $w(\mathbf{x}) = p_\mathbf{x}(\mathbf{x})/p_\mu(\mu(\mathbf{x}))$
     \STATE $w_\mathit{GMM}(\mathbf{x})$ $\leftarrow$ Fit GMM to $w(\mathbf{x})$

\end{ALC@g}
\algorithmicend~\algorithmicdo\
   \RETURN $a(\mathbf{x}; w_\mathit{GMM}(\mathbf{x}))$ and gradients in analytic form \end{algorithmic}
\end{algorithm}

\begin{table}[tbhp]
{\footnotesize
  \caption{Summary of the acquisition functions considered in this work.}\label{tab:1}
\begin{center}
\begin{tabular}{|c|c|c|} \hline
Acquisition Function & Equation & Rule  \\ \hline
US and variants		& \cref{eq:25}, \cref{eq:35} 	& maximize \\
IVR and variants	    	& \cref{eq:27}, \cref{eq:29}, \cref{eq:33} 	& maximize  \\
MI					& \cref{eq:212}				& maximize  \\
Q     					& \cref{eq:32}	& minimize  \\ 
L					& \cref{eq:34}	& minimize \\
\hline 
\end{tabular}
\end{center}
}
\end{table}

\section{Results}
\label{sec:4}

\subsection{Experimental Protocol}
\label{sec:41}

To demonstrate the benefits of the likelihood ratio in BED, we perform a series of numerical experiments with the acquisition functions introduced in \cref{sec:3}.  Specifically, we consider US, US-LW, IVR, IVR-IW, and the Q criterion.  (We do not include MI in this list because of the limitations discussed in \cref{sec:23}.)  For US-LW, we denote by US-LW the case in which $w(\mathbf{x})$ is approximated by a GMM, and US-LW\textsubscript{raw} the case in which it is not.  For the Q criterion, we use the IVR-LW formulation in \cref{eq:33}.   We compare the Bayesian sequential algorithm with Latin hypercube sampling (LHS), a non-iterative technique widely used in the experimental-design community \cite{mckay1979comparison}.  Compared to \cref{alg:1}, LHS has a higher computational cost as the number of black-box evaluations for LHS scales like the square of the number of iterations.

Our implementation is based on the package \texttt{gpsearch} available on GitHub\footnote{\url{https://github.com/ablancha/gpsearch}}.  For each example considered, we run 100 Bayesian experiments, each differing in the choice of initial points.  The algorithm is initialized with $d+1$ points sampled from an LHS design, as in \cite{hutter2011sequential}.  As discussed in \cref{sec:22}, we use the RBF kernel with ARD for the GP model.  Unless otherwise indicated, the noise variance $\sigma_\varepsilon^2$ is treated as a hyper-parameter and, as such, inferred from data.  The number of Gaussian mixtures used in \cref{eq:37} is kept constant throughout the search.  We use ``full'' covariance matrices in \cref{eq:37}.    

%

To evaluate performance, we report the median for the log-pdf error
\begin{equation}
e(n) = \int |\log p_{\mu_n}(y) - \log p_f(y)| \, \mathrm{d} y,
\label{eq:42}
\end{equation}
which quantifies the discrepancy between the pdf of the posterior mean at iteration $n$ ($\mu_n$) and the pdf of the true map $f$, both of which being computed by Monte Carlo sampling of the input space.  \Cref{eq:42} is used to assess the goodness of the model \textit{globally} by comparing the statistics of the predicted output to the ground truth, with the logarithms placing extra emphasis on the tails of the pdfs.  We note that \cref{eq:42} is more stringent than the root-mean-squared prediction error, another ``global'' error metric commonly used in BED \cite{beck2016sequential}, as the latter only measures the second moment of the differences between predicted values and actual values.  For \cref{alg:1}, we report the cumulative minimum, $\min_{k\in[0,n]} e(k)$, as is common for iterative methods (e.g., \cite{wang2017max}).

\subsection{Quantifying Rare Events in a Stochastic Oscillator}
\label{sec:42}

\subsubsection{Problem Formulation} 

We consider the stochastic oscillator of Mohamad and Sapsis \cite{mohamad2018sequential},
\begin{equation}
\ddot{u} + \delta \dot{u} + F(u) = \xi(t),
\label{eq:43}
\end{equation}
where $u(t) \in \mathbb{R}$ is the state variable, overdot denotes differentiation with respect to the time variable $t$, $\xi(t)$ is a stationary stochastic process with correlation function $\sigma_\xi^2 \exp[-\tau^2/(2\ell_{\xi}^2)]$, and $F$ is a nonlinear restoring force given by
\begin{equation}
F(u) = \begin{cases} 
	\alpha u,		& \text{for~} 0 \leq |u| \leq u_1, 		\\
	\alpha u_1,	& \text{for~}u_1 \leq |u| \leq u_2,	\\
	\alpha u_1 + \beta(u-u_2)^3,	& \text{for~}u_2 \leq |u|.
\end{cases}
\label{eq:44}
\end{equation}
As in \cite{mohamad2018sequential}, we parametrize the stochastic excitation $\xi(t)$ by a finite number of random variables using the Karhunen--Lo{\`e}ve expansion
\begin{equation}
\xi(t) \approx \mathbf{x} \mathbf{\Phi}(t), \quad t\in[0,T],
\label{eq:45}
\end{equation}
where $\mathbf{x} \in \mathbb{R}^m$ is a vector of coefficients normally distributed with zero mean and diagonal covariance matrix $\mathbf{\Lambda}$, and $\{\mathbf{\Lambda}, \mathbf{\Phi}(t)\}$ contains the first $m$ eigenpairs of the correlation matrix.  We use\footnote{Mohamad and Sapsis \cite{mohamad2018sequential} reported using $\sigma_\xi^2=4$ and $\ell_{\xi}=0.1$, while in fact they used $\sigma_\xi^2=0.1$ and $\ell_{\xi}=4$.} parameters $\delta=1.5$, $\alpha=1$, $\beta=0.1$, $u_1=0.5$, $u_2=1.5$, $\sigma_\xi^2=0.1$, $\ell_{\xi}=4$, and $T=25$.  The quantity of interest is taken to be the mean value of $u(t)$ over the interval $[0,T]$:
\begin{equation}
f(\mathbf{x}) = \frac{1}{T} \int_0^T u(t; \mathbf{x})\,\mathrm{d}t,
\label{eq:46}
\end{equation}
where we have made explicit the dependence of the response $u(t)$ on the random vector $\mathbf{x}$. 

\subsubsection{A Simple Illustration} 
\begin{figure}[tbhp]
\centering 
\subfloat[]{\label{fig:2a}\includegraphics[width=2.05in]{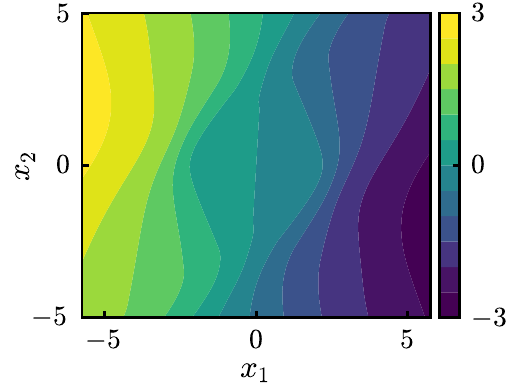}}
\subfloat[]{\label{fig:2b}\includegraphics[width=2.05in]{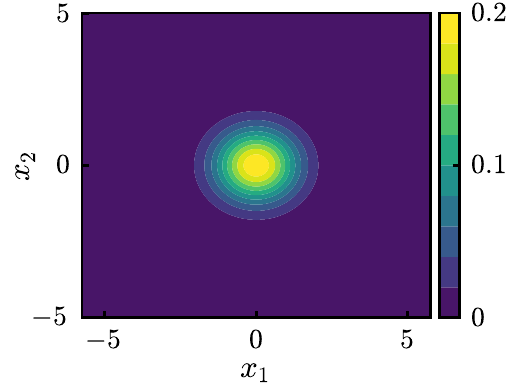}} ~~
\subfloat[]{\label{fig:2c}\includegraphics[width=1.8in]{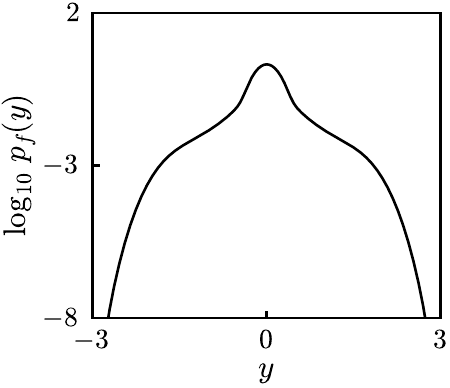}}
\caption{For the stochastic oscillator \cref{eq:43}, (a) contour plot of the objective function \cref{eq:46}, (b) contour plot of the input pdf $p_\mathbf{x}(\mathbf{x})$, and (c) pdf of the objective function conditioned on the input.}
\label{fig:2}
\end{figure}

To facilitate visualization of the algorithm's progress in the input space, we first consider a two-dimensional truncation ($m=2$) of \cref{eq:45}.   For the search space, we require that $\mathbf{x}$ lie no more than six standard deviations away from its mean in each direction.  For these parameters, \cref{fig:2} shows that the regions associated with large output values have low probability of occurrence.  This, combined with the strongly nonlinear nature of the stochastic oscillator \cref{eq:43}, gives rise to a heavy-tailed distribution for the output (\cref{fig:2c}).

\begin{figure}[!ht]
\centering 
\subfloat[]{\label{fig:3a}\includegraphics[width=1.8in]{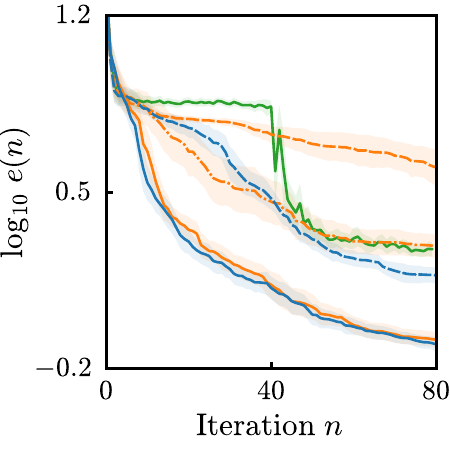}} \qquad
\subfloat[]{\label{fig:3b}\includegraphics[width=1.8in]{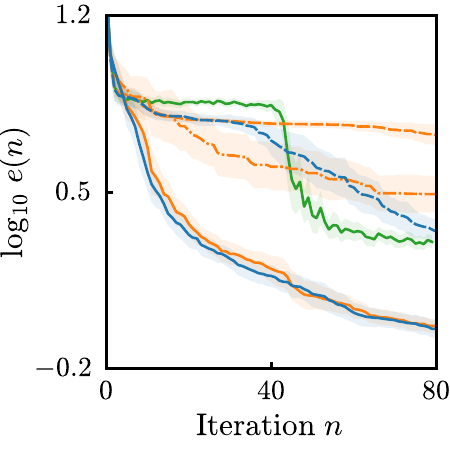}} \qquad
\subfloat[]{\label{fig:3c}\includegraphics[width=1.8in]{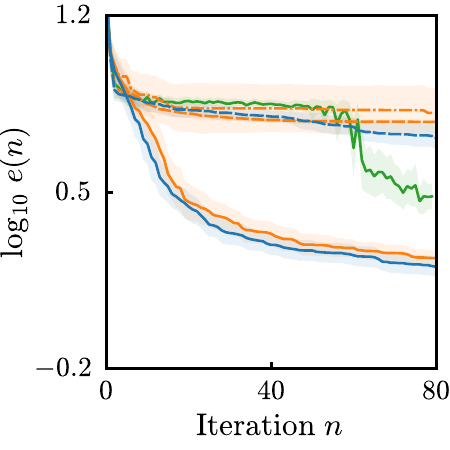}} 
\caption{For the stochastic oscillator \cref{eq:43} with $m=2$, performance of US (\usline), US-LW (\uslwline), IVR (\ivrline), IVR-IW (\ivriwline), IVR-LW (\ivrlwline) and LHS (\lhsline) for $n_\textit{GMM}=2$ and (a) $\sigma_\varepsilon^2 = 0$, (b) $\sigma_\varepsilon^2 = 10^{-3}$, and (c) $\sigma_\varepsilon^2 = 10^{-2}$. The error bands indicate one half of the median absolute deviation.}
\label{fig:3}
\end{figure}

\begin{figure}[!ht]
\centering 
\subfloat[]{\label{fig:4a}\includegraphics[width=1.8in]{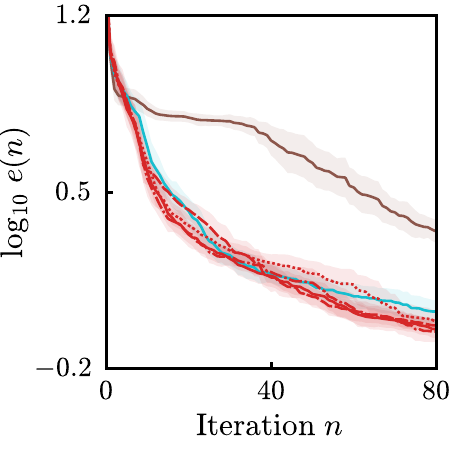}} \qquad
\subfloat[]{\label{fig:4b}\includegraphics[width=1.8in]{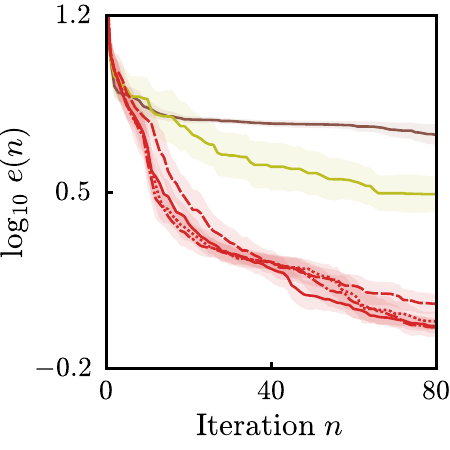}}
\caption{For the stochastic oscillator \cref{eq:43} with $m=2$ and $\sigma_\varepsilon^2 = 10^{-3}$, (a) performance of US (\solidbrown), US-LW\textsubscript{raw} (\solidcyan), and US-LW with $n_\textit{GMM}=1$ (\dashedred), $n_\textit{GMM}=2$ (\solidred), $n_\textit{GMM}=4$ (\dashdottedred), and $n_\textit{GMM}=6$ (\dottedred); and (b) performance of IVR (\solidbrown), IVR-IW (\solidyellow), and IVR-LW with $n_\textit{GMM}=1$ (\dashedred), $n_\textit{GMM}=2$ (\solidred), $n_\textit{GMM}=4$ (\dashdottedred), and $n_\textit{GMM}=6$ (\dottedred).  The error bands indicate one half of the median absolute deviation.}
\label{fig:4}
\end{figure}

For $n_\textit{GMM}=2$, \cref{fig:3} shows that the LW acquisition functions deliver better performance than their unweighted counterparts, independently of the noise level.  For $\sigma_\varepsilon^2 = 10^{-3}$, \cref{fig:4} provides evidence that the LW acquisition functions are not very sensitive to the number of Gaussian mixtures used in the approximation of $w(\mathbf{x})$.  In particular, \cref{fig:4a} shows that there is virtually no difference in performance between US-LW and US-LW\textsubscript{raw}, showing that the GMM approximation does not impede convergence of the algorithm.  Overall, \cref{fig:3,fig:4} show that the best-performing acquisition functions are US-LW and IVR-LW (both offering nearly identical performance), thus cementing the utility of the likelihood ratio in BED.

\begin{figure}[!t]
\centering 
\setcounter{subfigure}{-1}
\subfloat{\label{fig:5a1}\includegraphics[width=2.05in]{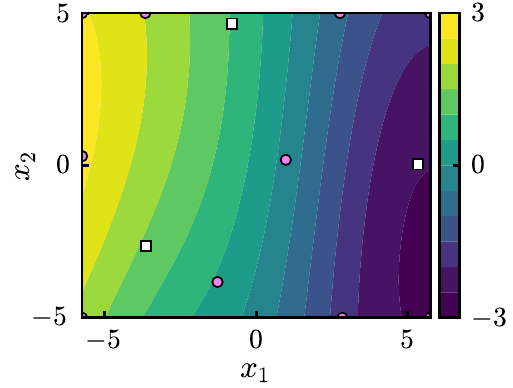}} 
\subfloat[]{\label{fig:5a2}\includegraphics[width=2.05in]{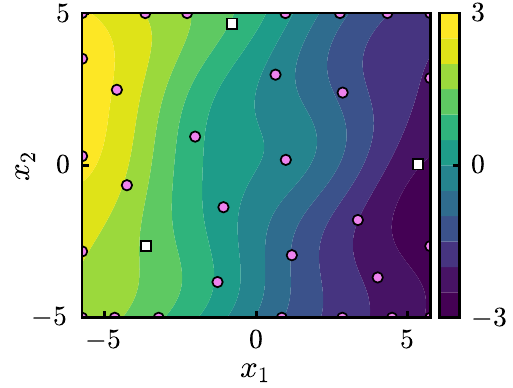}} 
\subfloat{\label{fig:5a3}\includegraphics[width=2.05in]{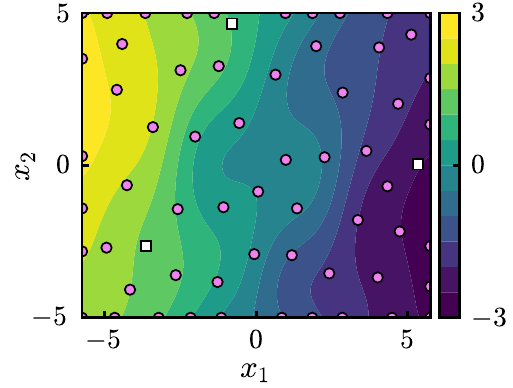}}
\vspace{-10pt}
\setcounter{subfigure}{0}
\subfloat{\label{fig:5ba}\includegraphics[width=2.05in]{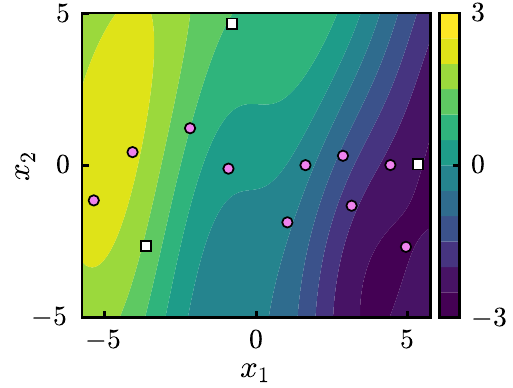}} 
\subfloat[]{\label{fig:5b2}\includegraphics[width=2.05in]{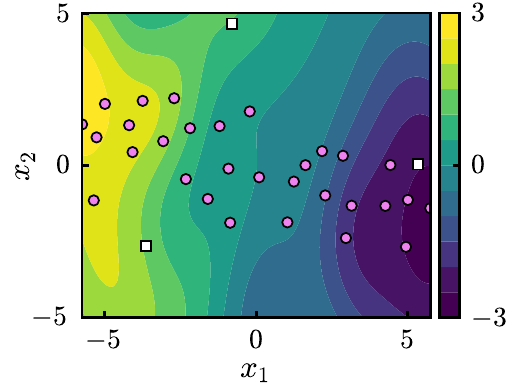}} 
\subfloat{\label{fig:5b3}\includegraphics[width=2.05in]{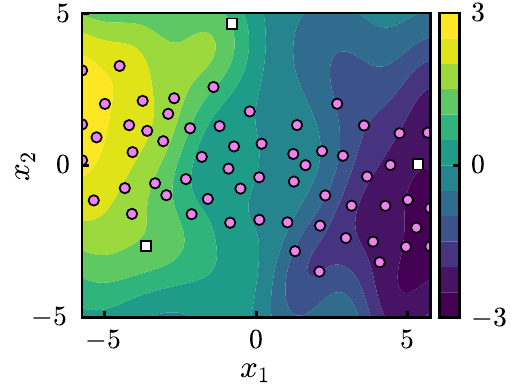}}
\vspace{-10pt}
\setcounter{subfigure}{1}
\subfloat{\label{fig:5c1}\includegraphics[width=2.05in]{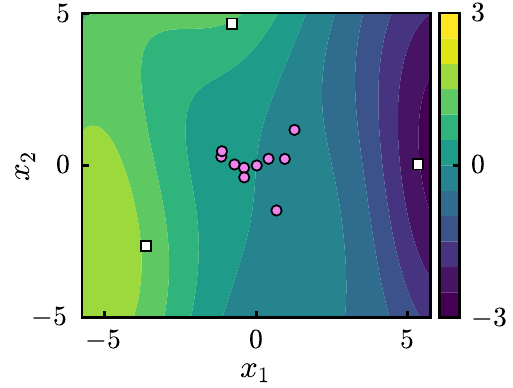}} 
\subfloat[]{\label{fig:5c2}\includegraphics[width=2.05in]{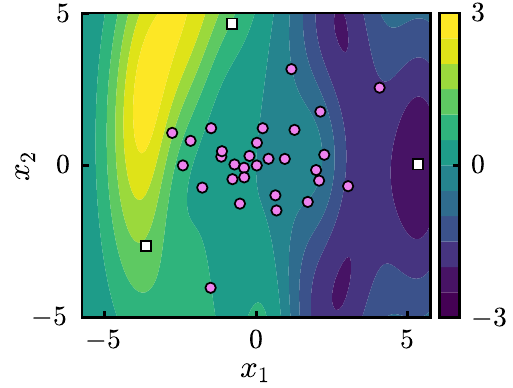}} 
\subfloat{\label{fig:5c3}\includegraphics[width=2.05in]{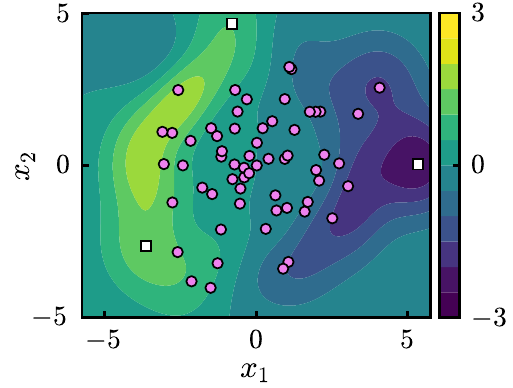}}
\vspace{-10pt}
\setcounter{subfigure}{2}
\subfloat{\label{fig:5d1}\includegraphics[width=2.05in]{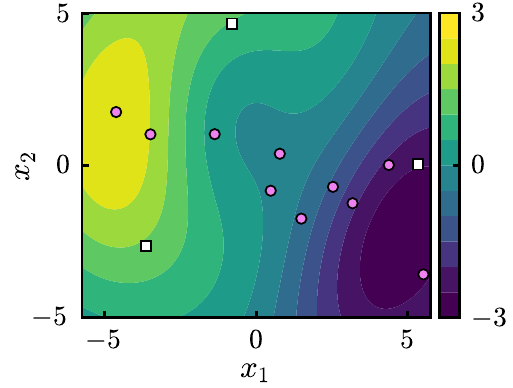}} 
\subfloat[]{\label{fig:5d2}\includegraphics[width=2.05in]{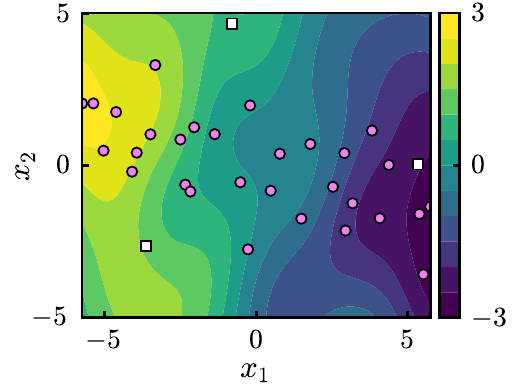}} 
\subfloat{\label{fig:5d3}\includegraphics[width=2.05in]{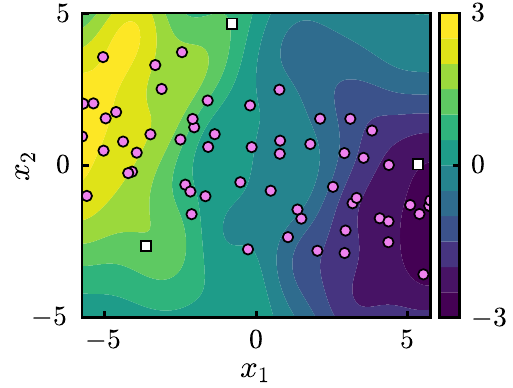}}

\caption{For the stochastic oscillator \cref{eq:43} with $m=2$, $\sigma_\varepsilon^2=0$ and $n_\textit{GMM}=2$, progression of the sampling algorithm for (a) US, (b) US-LW, (c) IVR-IW, and (d) IVR-LW after 10 iterations (left), 30 iterations (center), and 60 iterations (right); the contours denote the posterior mean of the GP model, the open squares the three initial LHS points, and the filled circles the points visited by the search algorithm.}
\label{fig:5}
\end{figure}

To explain the success of the LW acquisition functions, we monitor the progression of the search algorithm for noiseless observations and $\sigma_\varepsilon^2$ set to zero in the GP model.  \cref{fig:5} compares the decisions made by US, US-LW, IVR-IW and IVR-LW after 10, 30 and 60 iterations.  As discussed in \cref{sec:23}, US attempts to reduce uncertainty with no regard for the input distribution or the output values, resulting in a relatively uniform ``carpeting'' of the input space (\cref{fig:5a2}).  \cref{fig:5a2} also illustrates the propensity of US to visit points on the boundaries of the domain.  In contrast, \cref{fig:5b2} shows that US-LW focuses its effort on a diagonal band, ignoring the regions of lesser significance above and below that band.  The resulting surrogate model is able to predict the statistics of the output much more accurately than with US.  Another benefit of the likelihood ratio is that US-LW does not share the excessive interest of US in boundary points.

\cref{fig:5d2} shows that IVR-LW follows a similar strategy to US-LW, avoiding the vicinity of the upper right and lower left corners of the domain.  Compared to US-LW, IVR-LW sometimes visits a point despite it being in close proximity with one or more samples collected previously.  This is likely due to the fact that IVR-LW takes the form of an integral over the input space, as opposed to US-LW which acts ``pointwise'' in the domain.  In fact, that behavior is also seen in IVR and IVR-IW.  For the latter, \cref{fig:5c2} shows that the algorithm focuses exclusively on the center region of the input space where $p_\mathbf{x}$ is large.  Thus, the less likely yet more extreme regions are not visited, resulting in the surrogate model not being able to capture the heavy tails of the output pdf.

\subsubsection{Comparison to the Method of Active Subspaces}

To solidify the claim that the likelihood ratio helps identify critical regions of the input space more efficiently than otherwise, we compare the performance of US-LW and IVR-LW to the method of active subspaces \cite{constantine2015active}.  

The method of active subspaces is a popular technique for dimensionality reduction which allows identification of the ``active'' directions in the space of inputs when the latter is high-dimensional and the input--output relationship is expensive to evaluate.  For an unknown black-box function $f$, the construction of the active subspace involves the following steps: 
\begin{enumerate}
\item Draw $M=\lfloor \alpha k \log(d) \rfloor$ independent samples $\{\mathbf{x}_i\}_{i=1}^M$ from the distribution $p_\mathbf{x}$;
\item For each $\mathbf{x}_i$, compute the gradients $\nabla f(\mathbf{x}_i)$ by finite differences or some other method;
\item Compute the singular value decomposition of the matrix $\mathbf{C} = \begin{bmatrix} \nabla f(\mathbf{x}_1), \dots, \nabla f(\mathbf{x}_M) \end{bmatrix}$.
\end{enumerate}
In the above, $k$ is the number of singular values to be resolved accurately, and $\alpha$ is an oversampling factor to be specified between two and ten.  The active subspace $\mathbf{W}$ of size $q$ is the subspace spanned by the leading $q$ eigenvectors of the matrix $\mathbf{C}$.  

Once the active subspace $\mathbf{W}$ has been determined, it is possible to construct a surrogate model on the active variables by evaluating the black box at a small number of points in the active subspace and fitting a regression surface to the resulting input--output pairs.  The surrogate model $\mu$ can then be used to make predictions at any point $\mathbf{x}$ in the original space:
\begin{equation}
f(\mathbf{x}) \approx \mu(\mathbf{W}^\mathsf{T} \mathbf{x}).
\end{equation}
Sampling in the active subspace can be done in a number of ways, e.g., uniformly, on a grid, according to an LHS design, or by using leftover data from the process of discovering the active subspace in Steps 1--3 above.

Unlike \cref{alg:1}, the method of active subspaces, in its most common manifestation, is not iterative.  If the gradients of $f$ are computed by first-order finite differences, then construction of the active subspace requires $\lfloor \alpha k \log(d)\rfloor (d+1)$ function evaluations.  For an active subspace of size $q$ and the LHS policy used earlier, the minimum number of function evaluations needed to construct the surrogate model climbs to $N=\lfloor \alpha k \log(d)\rfloor (d+1) + (q+1)$.   Even for moderate dimensions, this number can be enormous; for $d=10$, $q=2$ and $\alpha=2$ (the smallest recommended value for the oversampling factor), we find that $N=102$, 201 and 300 for $k=2$, 4 and 6, respectively.  

\begin{figure}[!ht]
\centering 
\subfloat[]{\label{fig:10a}\includegraphics[width=1.8in]{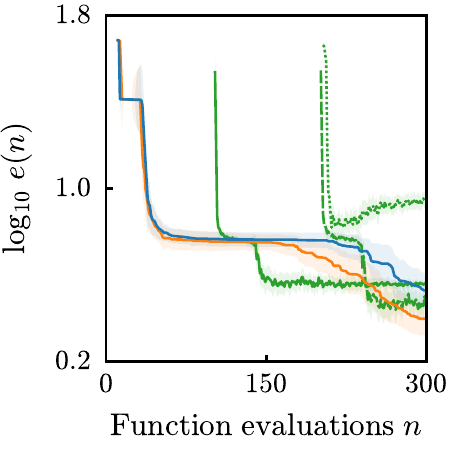}} \qquad
\subfloat[]{\label{fig:10b}\includegraphics[width=1.8in]{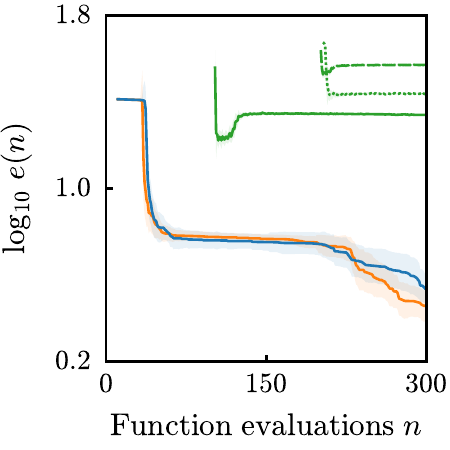}}
\caption{For the stochastic oscillator \cref{eq:43} with $m=10$, performance of US-LW (\uslwline), IVR-LW (\ivrlwline), and LHS of the active subspace with $k=2$ and $q=2$ (\lhsline), $k=4$ and $q=2$ (\lhslinedashed) and $k=4$ and $q=4$ (\lhslinedotted) for (a) $\sigma_\varepsilon^2 = 0$ and (b) $\sigma_\varepsilon^2 = 10^{-3}$. The error bands indicate one half of the median absolute deviation. }
\label{fig:10}
\end{figure}

We consider a ten-dimensional truncation ($m=10$) of \cref{eq:45} with objective function \cref{eq:46}.  \cref{fig:10} shows the log-pdf error as a function of total number of function evaluations for noiseless ($\sigma_\varepsilon^2=0$) and noisy  ($\sigma_\varepsilon^2=10^{-3}$) observations.  For noiseless observations, \cref{fig:10a} shows that \cref{alg:1} outperforms the method of active subspace in the two important limits of very small and very large $n$.  In between there is a range in which the method of active subspace is able to keep up with \cref{alg:1} but only for certain values of $q$ and $k$.  For noisy observations, \cref{fig:10b} shows that \cref{alg:1} outperforms the method of active subspace across the board and by a substantial amount.

When equipped with an LW acquisition function, \cref{alg:1} has several advantages compared to the method of active subspaces, especially in situations where the black-box function has the ability to generate rare events.  In particular, we note that:
\begin{enumerate}
\item The method of active subspaces works best when an accurate mechanism for gradient computation is readily available.  In many applications related to design and optimization, however, the quantity of interest is a black box, with nothing known about its inner workings, and surely no way of computing its gradients (\cref{sec:21}).  The finite-difference approximation recommended in \cite{constantine2015active} requires a much larger number of functions evaluations than the sequential algorithm \cref{alg:1}, and is not robust to noise, as made clear in \cref{fig:10b}. 
\item The active subspace is constructed by identifying the principal directions in a large number of samples drawn from the input distribution $p_\mathbf{x}$.  In the context of rare-event quantification, this approach is problematic because it is often the case that the regions of the input space responsible for rare events are associated with low probability of occurrence.  In addition, the active subspace is a linear subspace and therefore fails to account for the nonlinear features of the black-box function which are so critical in the generation of extreme events.
\item \cref{alg:1} is asymptotically correct in the sense that convergence of the statistics is guaranteed in the limit of many observations because the search is performed in the original space.  In the method of active subspace, however, the surrogate model only acts on the active variables which are the result of truncating the original space to a smaller number of dimensions.  It is not possible to recover from this truncation error by sampling the active subspace infinitely many times (\cref{fig:10}).
\end{enumerate}

\subsection{Quantifying parametric uncertainty in a hydrological model}
\label{sec:44}

For another high-dimensional example, we consider the hydrological model of Harper and Gupta \cite{harper1983sensitivity} which expresses the flow rate of water through a borehole as
\begin{equation}
y_0 = \frac{2 \pi T_u (H_u-H_l)}{\ln(r/r_w)}\left[ 1 + \frac{2 L T_u}{\ln(r/r_w) r_w^2 K_w} + \frac{T_u}{T_l} \right]\!,
\label{eq:47}
\end{equation}
where the definitions, ranges, and distributions of the eight input parameters are given in \cref{tab:2}.  (In the numerical experiments, the input space is rescaled to the unit hypercube in order to facilitate training of the GP hyper-parameters.)  The borehole function has been widely used as a test case for validation and benchmarking of active-learning computer codes \cite{morris1993bayesian,joseph2008blind,gramacy2012gaussian,xiong2013sequential}.  Our goal is to use Algorithm \ref{alg:1} to compute the pdf of $y_0$ efficiently given the uncertainty in the input parameters.  Given the distributions of the inputs and the strongly nonlinear nature of \cref{eq:47}, the true pdf of $y_0$ is heavy-tailed.  This makes the borehole function a good test case for the LW acquisition functions.

\begin{table}[tbhp]
{\footnotesize
  \caption{Parameters for the borehole function \cref{eq:47}.  The normal distribution for $r_w$ has mean 0.1 and standard deviation 0.0161812; the lognormal distribution for $r$ has central value 7.71 and uncertainty factor 1.0056.}\label{tab:2}
\begin{center}
\begin{tabular}{|c|c|c|c|} \hline
Parameter & Definition 								&  Range			& Distribution 	\\ \hline
$r_w$	& Radius of borehole (m)						&  $[0.05, 0.15]$	& Normal \\ 
$r$		& Radius of influence (m)						&  $[100, 50000]$	& Lognormal \\ 
$T_u$	& Transmissivity of upper aquifer (m\textsuperscript{2}/yr)	& $[63070, 115600]$	& Uniform   \\
$H_u$	& Potentiometric head of upper aquifer (m)		& $[990, 1110]$		& Uniform \\
$T_l$	& Transmissivity of lower aquifer (m\textsuperscript{2}/yr)	& $[63.1, 116]$	& Uniform	 \\
$H_l$	& Potentiometric head of lower aquifer (m)		& $[700, 820]$		& Uniform \\
$L$		& Length of borehole (m)						& $[1120, 1680]$	& Uniform \\
$K_w$	& Hydraulic conductivity of borehole (m/yr)		& $[9855, 12045]$	& Uniform \\
\hline \end{tabular}
\end{center}
}
\end{table}

\begin{figure}[!ht]
\centering 
\subfloat[]{\label{fig:6a}\includegraphics[width=1.8in]{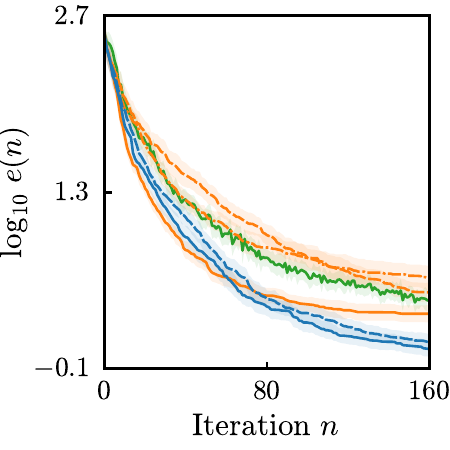}} \qquad
\subfloat[]{\label{fig:6b}\includegraphics[width=1.8in]{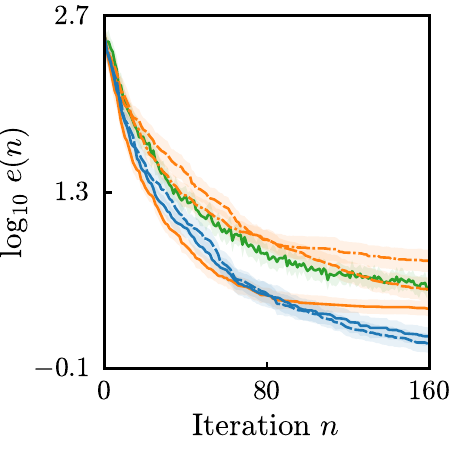}} \qquad
\subfloat[]{\label{fig:6c}\includegraphics[width=1.8in]{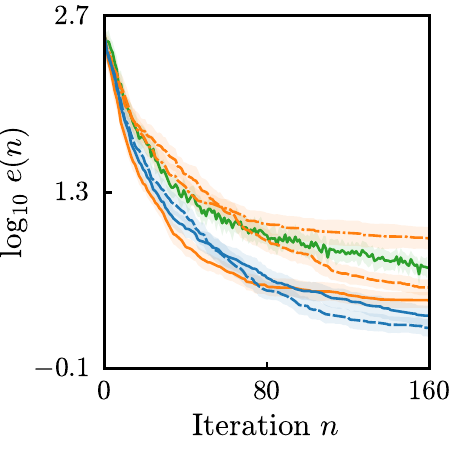}} 
\caption{For the borehole function \cref{eq:47}, performance of US (\usline), US-LW (\uslwline), IVR (\ivrline), IVR-IW (\ivriwline), IVR-LW (\ivrlwline) and LHS (\lhsline) for $n_\textit{GMM}=2$ and (a) $\sigma_\varepsilon^2 = 0$, (b) $\sigma_\varepsilon^2 = 10^{-3}$, and (c) $\sigma_\varepsilon^2 = 10^{-2}$. The error bands indicate one half of the median absolute deviation.}
\label{fig:6}
\end{figure}

For $n_\textit{GMM}=2$ and three values of the noise variance $\sigma_\varepsilon^2$, \cref{fig:6} shows that IVR-LW leads to faster convergence than IVR-IW and IVR.  For US and US-LW, however, the gains are not as substantial and the performance of US-LW is on par with that of US.  We conjecture that this is due to the fact that six of the eight input parameters follow uniform distributions, which might have the effect of diluting some of the benefits of the likelihood ratio in US-LW.  (It should be clear that when $p_\mathbf{x}$ is uniform, the only effect of the likelihood ratio is through the density of the posterior mean $p_\mu$.)  

\subsection{Mapping the geography of extreme events in dynamical systems}
\label{sec:45}

\subsubsection{Problem Formulation} 
\label{sec:451}

As a final example, we consider the problem of predicting the occurrence of extreme events in dynamical systems.  The central issue is to identify \textit{precursors}, i.e., those states of the system which are most likely to lead to an extreme event in the near future.  Searching for precursors is no easy task because extreme events often arise in highly complex dynamical systems, which adds to the issue of low frequency of occurrence.  We propose to use Algorithm \ref{alg:1} to parsimoniously probe the state space of the system and thus identify ``dangerous'' regions using as little data as possible.  

Formally, the dynamical system is treated as a black box which assigns to any point in the phase space $\mathbf{x}$ a measure of dangerousness, e.g.,
\begin{equation}
f(\mathbf{x}) =   \max_{t\in[0, \tau]} g(S_t(\mathbf{x})).
\label{eq:20}
\end{equation}
Here, $t$ denotes the time variable, $S_t$ the flow map of the system (i.e., the dynamics of the black box), $g : \mathbb{R}^d \longrightarrow \mathbb{R} $ the observable of interest, and $\tau$ the time horizon over which prediction is to be performed.  In words, \cref{eq:20} records the maximum value attained by the observable $g$ during the time interval $[0, \tau]$ given initial condition $\mathbf{x}$, and in a way defines a ``danger map'' for the dynamical system.  The role of Algorithm \ref{alg:1} is to search for those $\mathbf{x}$ that give rise to large values of $f$ indicative of an extreme event occurring within the next $\tau$ time units.  


In practice, not the whole phase space is explored by the algorithm because a) this would allow sampling of unrealistic states whose probability of being realized ``in the wild'' is essentially nil, and b) the dimension of the phase space can be unfathomably large (e.g., when $\mathbf{x}$ arises from the discretization of a partial differential equation).  To address the latter point, we adopt the approach of Farazmand and Sapsis \cite{farazmand2017variational} whereby extreme events are viewed as excursions from a ``background'' attractor for which a lower-dimensional representation can be constructed by principal component analysis (PCA).  

The advantage of PCA is that it approximates the statistics of the background attractor as a multivariate Gaussian distribution, and as a result the PCA subspace comes equipped with a Gaussian prior, $p_\mathbf{x}(\mathbf{x}) = \mathcal{N}(\mathbf{x};\mathbf{0}, \mathbf{\Lambda})$, with $\mathbf{\Lambda}$ a diagonal matrix containing the PCA eigenvalues.  PCA thus provides a mathematically consistent mechanism for generating random samples on the attractor.  To eliminate the possibility of drawing exotic samples, the search space is limited to a rectangular cuboid whose $i$th edge has length $a \sqrt{\lambda_i}$, where $a$ is typically of order one. This is slightly more conservative than the ellipsoid of Blonigan et al. \cite{blonigan2019extreme}.  

If the statistics of the attractor are strongly non-Gaussian, then one can use any other classical method for nonlinear dimensionality reduction and manifold learning.  For example, independent component analysis (ICA) provides an approximation of the attractor as a sum of statistically independent non-Gaussian components \cite{hyvarinen2013independent}.  Another possibility is to use embedding algorithms \cite{tenenbaum2000global,roweis2000nonlinear,zhang2004principal,maaten2008visualizing} and extract a prior $p_\mathbf{x}$ for the learned coordinates by kernel density estimation.  The advantage of a data-driven approach such as those mentioned here is that the data need not contain a single extreme event, as it is merely used to construct a finite-dimensional representation for the \textit{core} of the attractor.

\subsubsection{Application to a Nonlinear Dynamical System} 
\label{sec:452}

\begin{figure}[ht!]
\centering
\includegraphics[height=1.7in]{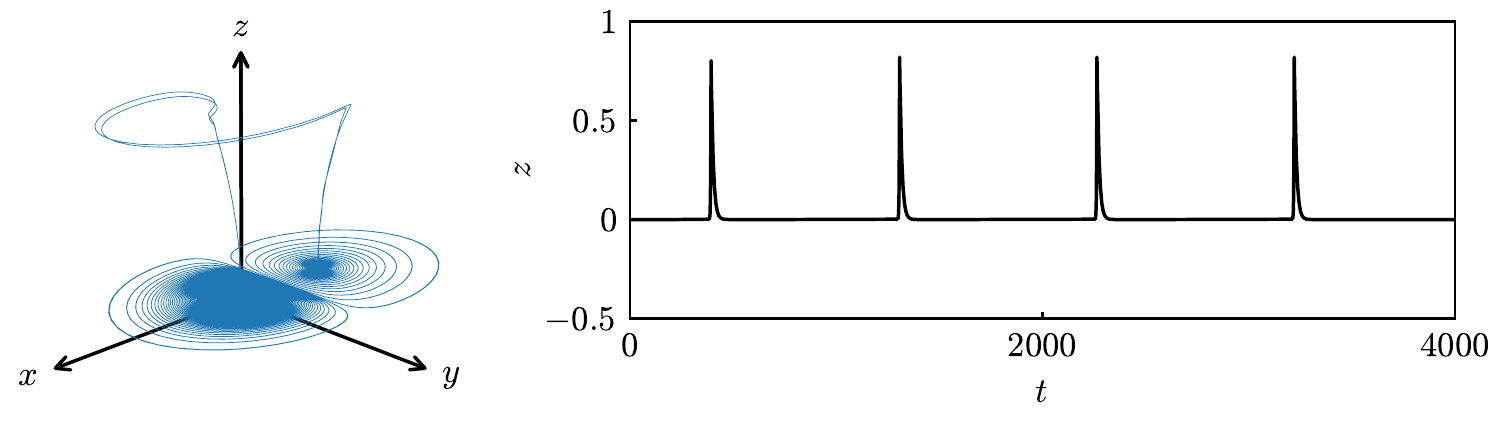}
\caption{For system \cref{eq:18}, trajectory in phase space (left) and time series for the $z$ coordinate (right).}
\label{fig:7}
\end{figure}

We consider a modified version of the dynamical system introduced by Farazmand and Sapsis \cite{farazmand2016dynamical}:
\begin{subequations}
\begin{gather}
\dot{x} = \alpha x + \omega y + \alpha x^6 + 2 \omega x y + 5 z^2, \label{eq:18a}\\
\dot{y} = -\omega x + \alpha y - \omega x^2 + 6 \alpha x y, \label{eq:18b}\\
\dot{z} = -\lambda z - (\lambda + \beta)x z, \label{eq:18c}
\end{gather}%
\label{eq:18}%
\end{subequations}%
with parameters $\alpha=0.02$, $\omega = 2\pi$, $\lambda=0.1$, and $\beta=0.7$.  \Cref{fig:7} shows that the system features successive ``cycles'' during which a trajectory initialized close to the origin spirals away towards the point $(-1,0,0)$, only to find itself swiftly repelled from the $z = 0$ plane.  After hovering about, the trajectory ultimately heads back to the origin and the cycle repeats itself.  Here, extreme events correspond to ``bursts'' in $z$ (\cref{fig:7}), and the dangerous region to the immediate vicinity of $(-1,0,0)$.

Our goal is to compute the danger map of the system for the observable $g=\mathbf{e}_z^\mathsf{T}$.  We use \cref{eq:20} as our measure of dangerousness and $\tau=50$ for the prediction horizon.  The background attractor is approximated with three principal components, with the leading two roughly spanning the $(x,y)$ plane.  We note that the approach to the fixed point $(-1,0,0)$ and the ensuing escape are significantly faster with \cref{eq:18} than in \cite{farazmand2016dynamical}, and as a result the core of the attractor is concentrated near the origin (\cref{fig:7}).  In this way, the dangerous region has a low yet non-negligible probability of being visited, making it difficult for US, IVR, and IVR-IW to identify that region.  For the dangerous region to be discoverable, we require that $\mathbf{x}$ lie no further than five PCA standard deviations in any direction (i.e., $a=5$).

\begin{figure}[ht!]
\centering
\subfloat[]{\label{fig:8a} \includegraphics[height=1.7in]{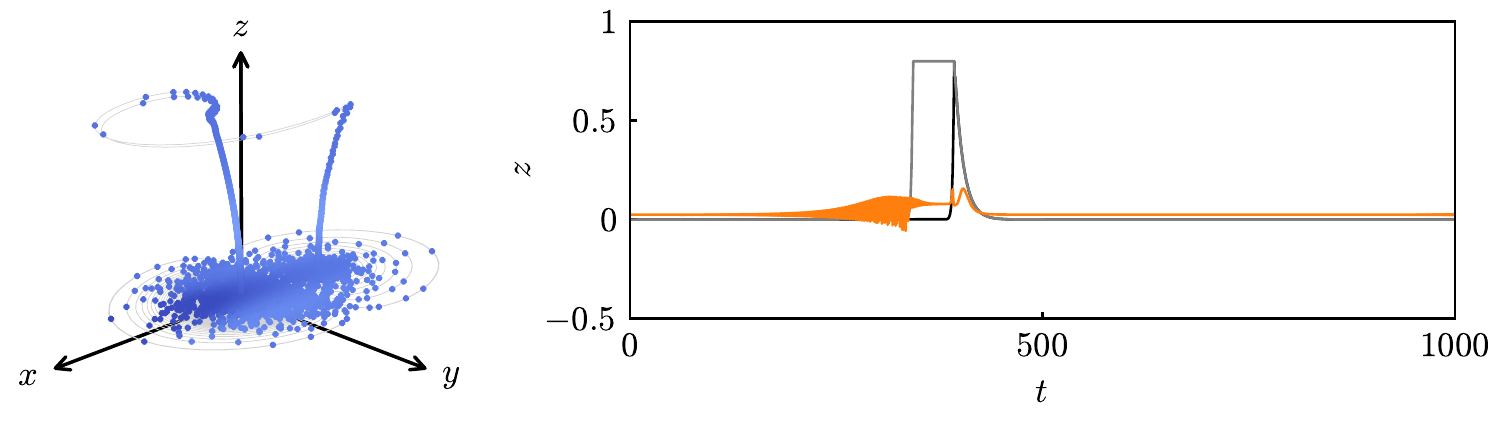}}  

\subfloat[]{\label{fig:8b} \includegraphics[height=1.7in]{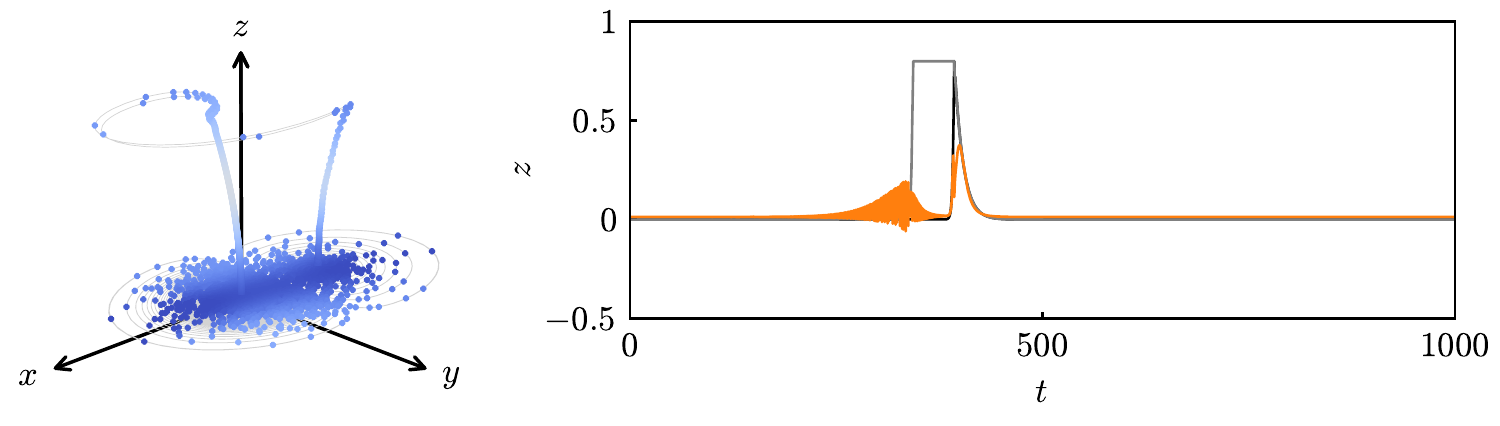}}  

\subfloat[]{\label{fig:8c} \includegraphics[height=1.7in]{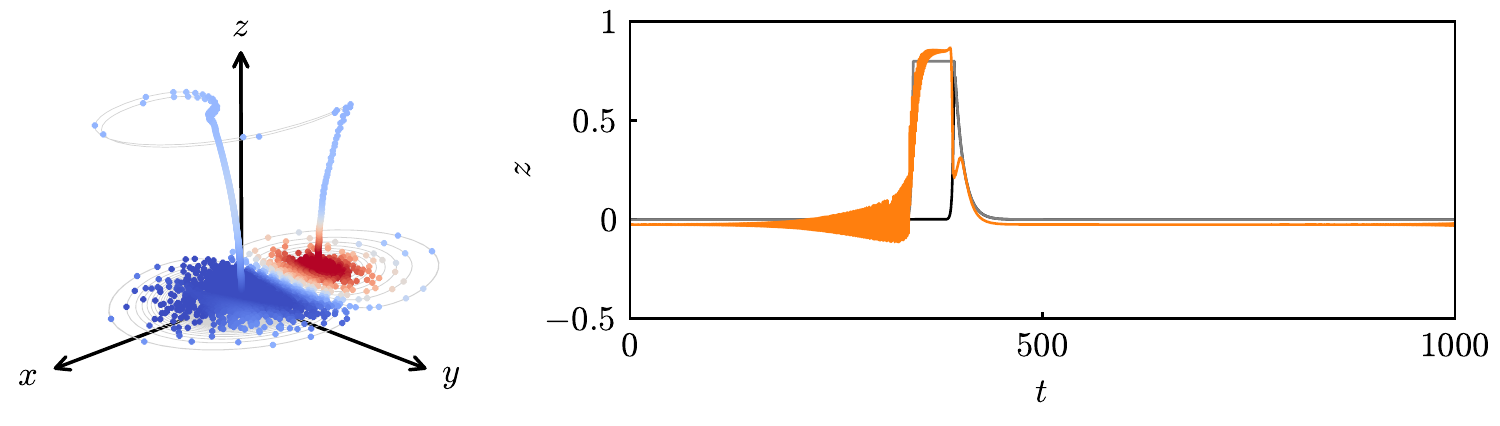}} 

\caption{Progression of the sequential sampling algorithm after (a) five iterations, (b) ten iterations, and (c) twenty iterations.  Left: trained danger map $\mu$ evaluated on test data with the false color scale ranging from 0 (blue) to 0.8 (red).  Right: time series for the observable $g$ (\blackline), the true danger map $f$ (\grayline), and the trained danger map $\mu$ (\orangeline).}
\label{fig:8}
\end{figure}

For $\sigma_\varepsilon^2=10^{-4}$ and IVR-LW with $n_\textit{GMM}=2$, \cref{fig:8} shows how the sampling algorithm progressively learns the danger map associated with the observable $g$.  \Cref{fig:8} shows 10,000 test points collected along a 2000-time-unit trajectory, each colored by dangerousness as predicted by the posterior mean $\mu(\mathbf{x})$.  (The test points should \textit{not} be confused with the optimized samples used to train the GP model.)  The cartography of extreme events becomes more accurate as the number of optimized samples grows.  After a few dozen iterations, the algorithm has correctly identified the region near $(-1,0,0)$ as being the most dangerous region on the attractor.  \Cref{fig:8} also shows how, if the test points are viewed as a continuously recorded stream of data, the posterior mean of the GP model can be used as an indicator to predict the occurrence of an extreme event in real time.

\begin{figure}[!ht]
\centering 
\subfloat[]{\label{fig:9a}\includegraphics[width=1.8in]{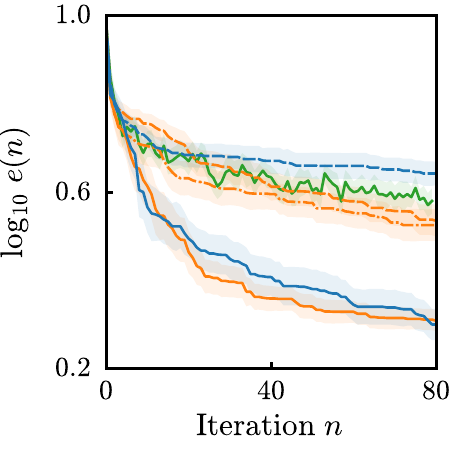}} \qquad
\subfloat[]{\label{fig:9b}\includegraphics[width=1.8in]{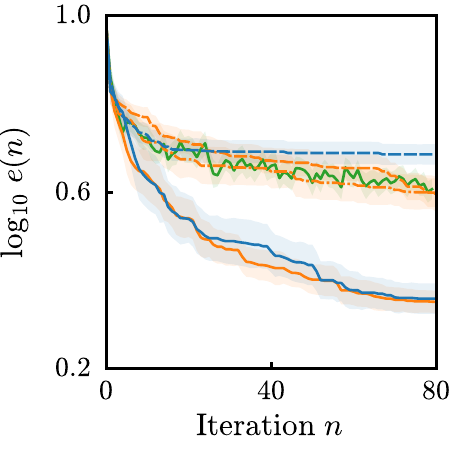}} \qquad
\subfloat[]{\label{fig:9c}\includegraphics[width=1.8in]{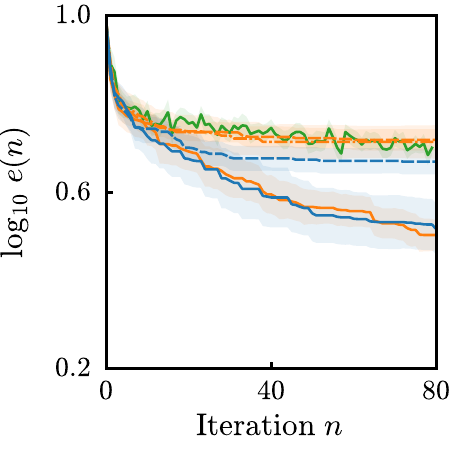}} 
\caption{For the dynamical system \cref{eq:18} featuring extreme events, performance of US (\usline), US-LW (\uslwline), IVR (\ivrline), IVR-IW (\ivriwline), IVR-LW (\ivrlwline) and LHS (\lhsline) for $n_\textit{GMM}=2$ and (a) $\sigma_\varepsilon^2 = 10^{-4}$, (b) $\sigma_\varepsilon^2 = 10^{-3}$, and (c) $\sigma_\varepsilon^2 = 10^{-2}$. The error bands indicate one quarter of the median absolute deviation.}
\label{fig:9}
\end{figure}

\Cref{fig:9} shows that the proposed algorithm is robust with respect to changes in $\sigma_\varepsilon^2$.  In each case shown in \cref{fig:9}, the LW acquisition functions deliver better performance than their unweighted cousins as well as LHS.  We note that use of the Gaussian prior $p_\mathbf{x}$ might lead to even greater gains in situations where the dynamical system features more than one dangerous region.


\begin{figure}[!ht]
\centering 
\subfloat[]{\label{fig:11a}\includegraphics[width=1.8in]{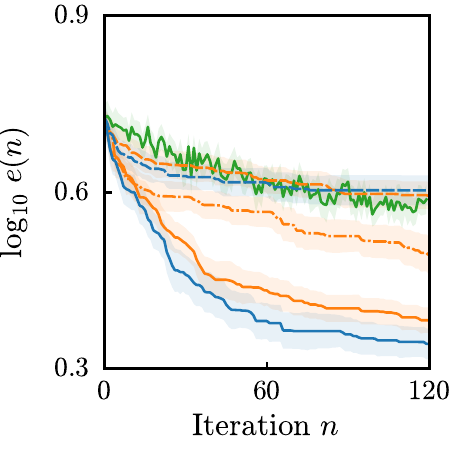}} \qquad
\subfloat[]{\label{fig:11b}\includegraphics[width=1.8in]{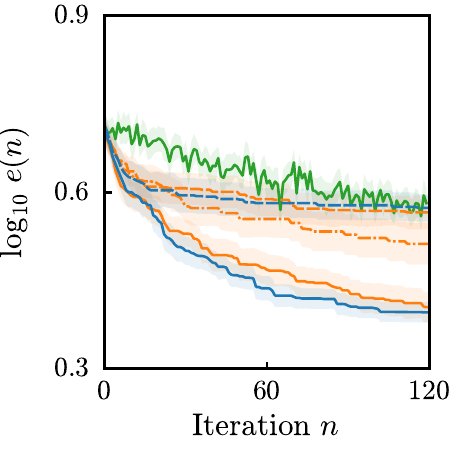}} \qquad
\subfloat[]{\label{fig:11c}\includegraphics[width=1.8in]{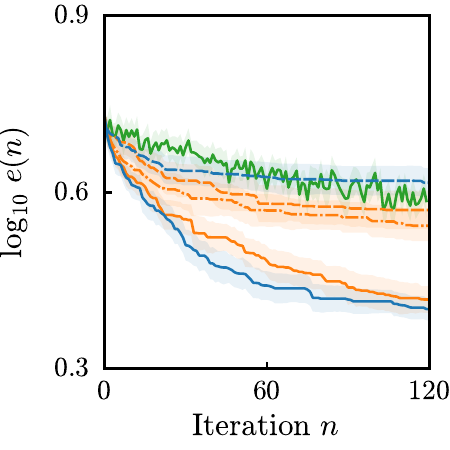}} 
\caption{For the dynamical system \cref{eq:18} featuring extreme events, performance of US (\usline), US-LW (\uslwline), IVR (\ivrline), IVR-IW (\ivriwline), IVR-LW (\ivrlwline) and LHS (\lhsline) for $n_\textit{GMM}=2$, $\sigma_\varepsilon^2 = 10^{-3}$, and (a) $d=10$, (b) $d=20$, and (c) $d=30$. The error bands indicate one quarter of the median absolute deviation.}
\label{fig:11}
\end{figure}

To investigate the effect of the dimension of the search space, we augment the three-dimensional PCA coordinate system with $d-3$ dummy dimensions that follow a standard normal distribution, similar to \cite{wang2016bayesian}.  (Along those directions, we use $a=0.2$.)  In this way, the dimension $d$ of the augmented space can be made arbitrarily large.  The added dimensions have no effect on the output of the black box, which puts the likelihood-weighted acquisition functions in a favorable position \cite{sapsis2020output}.  \Cref{fig:11} demonstrates that the benefits provided by the likelihood ratio carry over to high-dimensional spaces.
  
\subsubsection{Benefits of the Sampling Algorithm for Extreme-Event Prediction}

We conclude this section with a discussion of the benefits offered by Algorithm \ref{alg:1} in the context of extreme-event prediction and quantification in dynamical systems.  The key issue is that the algorithm directly learns the danger map $f$, opening many more doors than a probabilistic approach that only approximates the statistics of the observable $g$.  

As discussed in \cref{sec:451}, the algorithm produces a danger map of the attractor where each point is assigned a measure of dangerousness.  The fact that the measure of dangerousness is continuous rather than categorical allows for \emph{quantification} of danger at any point on the attractor.  For a given $\mathbf{x}$, the algorithm can assert not only whether an extreme event is about to happen, but, more importantly, the \emph{severity} of that event as recorded by the observable $g$.  The danger map thus provides the complete geography of extreme events on the attractor.  

The danger map also allows computation of precursors for each dangerous region (i.e., representative states for what danger looks like) by clustering states for which $\mu$ exceeds a certain threshold.  If more than one dangerous region exists, one can use multiple clusters and compute a precursor for each.  That the danger map can be evaluated at any point in the phase space implies that the search for precursors is not limited to a single precursor, nor is it sensitive to where the search is initiated, unlike in Farazmand and Sapsis \cite{farazmand2017variational}.  The proposed approach also requires no heavy machinery like the adjoint-state method used in \cite{farazmand2017variational}.

If in an experiment or application, it is possible to continuously record the state of the black box in the form of a data stream, then the danger map can be evaluated ``online'' every time a new point is received.  The resulting indicator ``senses'' the imminence of an extreme event \emph{in real time}. Real-time prediction is important from the standpoint of controlling extreme events, a task that often requires a good a priori understanding of the physical mechanisms responsible for the intermittent bursts \cite{farazmand2019closed}. Our approach, being agnostic to the details of the black-box dynamics, eliminates this requirement.

The algorithm is only limited by the topology of the dangerous regions on the attractor.  Situations in which these regions are fractal or riddled are likely to be problematic from the standpoint of making accurate predictions, as GP regression is essentially a linear smoother \cite{rasmussen2006gaussian}.  A closely-related issue is that of large Lyapunov exponents, giving rise to situations in which nearby trajectories diverge from each other exponentially rapidly.  If divergence happens within the prediction interval $[0, \tau]$, then our algorithm is likely to struggle because infinitesimally close input values will lead to large variations in output values.

\section{Conclusions}
\label{sec:5}

We have investigated the effect of including the likelihood ratio in several acquisition functions traditionally used in Bayesian experimental design of black-box functions.  The likelihood ratio assigns to each sample a measure of relevance that weighs how often that sample is likely to appear ``in the wild'' relative to the expected effect of that sample on the output value of the objective function. Compared to other information-based acquisition functions, the likelihood-weighted acquisition functions are tractable in high dimensions and computationally less complex.  We have found that the likelihood ratio accelerates convergence of the sequential algorithm in a number of examples related to uncertainty quantification and rare-event prediction.  The question of whether gains of similar proportions might be achieved in Bayesian optimization (where the focus is on learning the \textit{minimum} of the objective function rather than the objective function itself or its statistics) is considered in \cite{blanchard2020bayesian}.

We leave the reader with a few words on possible improvements for the algorithm.  First, our approach is not to be thought of as a way out of the no-free-lunch theorem \cite{wolpert1997no}.  In fact, our approach is expected to provide an advantage in situations where a sequential Bayesian algorithm is appropriate and the events of interest are sufficiently rare and extreme.  In other situations, the proposed approach may provide no advantage at all.  Second, while our approach has produced very encouraging results in high-dimensional spaces, improvements can be made towards further alleviating the curse of dimensionality, including leveraging information about the topology or curvature of the search space \cite{jaquier2020high}.

\section*{Acknowledgments}
The authors acknowledge support from the Army Research Office (Grant No. W911NF-17-1-0306), the MIT Doherty Career Development Chair, and the 2020 MathWorks Faculty Research Innovation Fellowship.

\bibliographystyle{siamplain}
\bibliography{bibl}

\appendix 

\section{Equivalence between \cref{eq:26} and \cref{eq:27}}
\label{app:1}

Let $\mathbf{x}$ be an arbitrary point in $\mathbb{R}^d$, and $\tilde{\mathbf{x}}$ a hypothetical ``ghost point'' with observation $\mu(\tilde{\mathbf{x}})$.  Assuming that the GP hyper-parameters are \textit{not} updated upon addition of the ghost pair $\{\tilde{\mathbf{x}}, \mu(\tilde{\mathbf{x}})\}$, we have
\begin{equation}
\sigma^2(\mathbf{x}) - \sigma^2(\mathbf{x}; \tilde{\mathbf{x}}) = - k(\mathbf{x}, \mathbf{X}) \mathbf{K}^{-1} k(\mathbf{X},\mathbf{x}) +  k(\mathbf{x}, \tilde{\mathbf{X}}) \tilde{\mathbf{K}}^{-1} k(\tilde{\mathbf{X}},\mathbf{x}),
\label{eq:aa3}
\end{equation}
where $\tilde{\mathbf{X}} = \mathbf{X} \cup \{\tilde{\mathbf{x}}\}$, and we have defined \begin{equation}
\tilde{\mathbf{K}} = 	\begin{bmatrix}  
			\mathbf{K} 		& k(\mathbf{X},\tilde{\mathbf{x}}) \\ 
			k(\tilde{\mathbf{x}},\mathbf{X}) 	& k(\tilde{\mathbf{x}},\tilde{\mathbf{x}}) 
					\end{bmatrix}.
\label{eq:ktilde}
\end{equation}
If we Cholesky-decompose $\mathbf{K}$ as $\mathbf{A}\mathbf{A}^\mathsf{T}$, then we can also Cholesky-decompose $\tilde{\mathbf{K}}$ as $ \tilde{\mathbf{A}}\tilde{\mathbf{A}}^\mathsf{T}$, with
\begin{equation}
 \tilde{\mathbf{A}} = 	\begin{bmatrix}  
		\mathbf{A} 				& \mathbf{0} \\ 
		k(\tilde{\mathbf{x}},\mathbf{X}) \mathbf{A}^{-\mathsf{T}} 	& \sigma(\tilde{\mathbf{x}}) 
		\end{bmatrix}.
\end{equation}
With this in hand, we compute
\begin{equation}
\tilde{\mathbf{A}}^{-1} = 	\frac{1}{\sigma(\tilde{\mathbf{x}})} \begin{bmatrix}  
			\sigma(\tilde{\mathbf{x}}) \mathbf{A}^{-1} 		& \mathbf{0}\\ 
			- k(\tilde{\mathbf{x}},\mathbf{X}) \mathbf{K}^{-1}   & 1
					\end{bmatrix},
\end{equation}
which leads to 
\begin{equation}
\tilde{\mathbf{K}}^{-1} = \frac{1}{\sigma^2(\tilde{\mathbf{x}})}
	\begin{bmatrix}  
		\sigma^2(\tilde{\mathbf{x}}) \mathbf{K}^{-1}  + \mathbf{K}^{-\mathsf{T}} k(\mathbf{X},\tilde{\mathbf{x}})k(\tilde{\mathbf{x}},\mathbf{X}) \mathbf{K}^{-1}				& -\mathbf{K}^{-\mathsf{T}} k(\mathbf{X},\tilde{\mathbf{x}}) \\ 
		-k(\tilde{\mathbf{x}},\mathbf{X}) \mathbf{K}^{-1} 		& 1
	\end{bmatrix}.
\label{eq:a7}
\end{equation}
We then substitute (\ref{eq:a7}) into (\ref{eq:aa3}), and after simplification obtain
\begin{equation}
\sigma^2(\mathbf{x})  - \sigma^2(\mathbf{x}; \tilde{\mathbf{x}})  = \frac{1}{\sigma^2(\tilde{\mathbf{x}})} \left[k(\tilde{\mathbf{x}},\mathbf{x}) -k(\tilde{\mathbf{x}},\mathbf{X}) \mathbf{K}^{-1} k(\mathbf{X},\mathbf{x})\right]^2,
\label{eq:noiless}
\end{equation}
which completes the proof.

In the above derivation it is assumed that the ghost pair $\{\tilde{\mathbf{x}}, \mu(\tilde{\mathbf{x}})\}$ is observed without noise corrupting the output.  This can be seen by evaluating the posterior variance at $\tilde{\mathbf{x}}$ after addition of the ghost pair:
\begin{equation}
\sigma^2(\tilde{\mathbf{x}}; \tilde{\mathbf{x}}) = \sigma^2(\tilde{\mathbf{x}}) - \frac{\mathrm{cov}^2(\tilde{\mathbf{x}}, \tilde{\mathbf{x}})}{\sigma^2(\tilde{\mathbf{x}})} = 0.
\end{equation}
Had we assumed a noise-corrupted ghost pair, the bottom right entry of the matrix $\tilde{\mathbf{K}}$ in \cref{eq:ktilde} would have read $k(\tilde{\mathbf{x}},\tilde{\mathbf{x}}) + \sigma_\varepsilon^2$, and the final result, 
\begin{equation}
\sigma^2(\mathbf{x})  - \sigma^2(\mathbf{x}; \tilde{\mathbf{x}})  = \frac{1}{\sigma^2(\tilde{\mathbf{x}})+ \sigma_\varepsilon^2} \left[k(\tilde{\mathbf{x}},\mathbf{x}) -k(\tilde{\mathbf{x}},\mathbf{X}) \mathbf{K}^{-1} k(\mathbf{X},\mathbf{x})\right]^2.
\label{eq:noisy}
\end{equation}
The only difference between \cref{eq:noisy} and \cref{eq:noiless} is the presence of the noise variance $\sigma_\varepsilon^2$ in the denominator of the term multiplying $\mathrm{cov}^2(\mathbf{x}, \tilde{\mathbf{x}})$.  

The assumption of a noiseless ghost point is appropriate for several reasons.  First, it is tantamount to a zeroth-order approximation of the noisy case in the limit of small $\sigma_\varepsilon^2$.  (To see this, perform a Taylor-series expansion of \cref{eq:noisy} in terms of $\sigma_\varepsilon^2$ and retain the leading-order term to recover \cref{eq:noiless}.)  This approximation is consistent with the general formulation of the problem in which it is assumed that the reason why each black-box query is expensive is precisely because the black box returns very accurate, nearly noiseless measurements.

Second, assuming a noiseless ghost pair encourages the algorithm to trust the GP model more than assuming a noisy ghost pair would.  Our numerical experiments suggest that the former approach can favorably affect the algorithm, especially in the early stages of the search when the algorithm is trying to determine the value of $\sigma_\varepsilon^2$ from a rather small dataset.  In the later stages of the search, or if the noise variance is \textit{fixed} beforehand, then this favorable effect is somewhat less pronounced.


\section{Analytical Expressions for IVR with RBF Kernel}
\label{app:2}

We first expand the formula for IVR using the GP expression for the posterior covariance:
\begin{subequations}
\begin{align}
\sigma^2(\mathbf{x})\, a_\textit{IVR}(\mathbf{x}) &= \int \left[k(\mathbf{x},\mathbf{x}') -k(\mathbf{x},\mathbf{X}) \mathbf{K}^{-1} k(\mathbf{X},\mathbf{x}')\right]^2  \, \mathrm{d}\mathbf{x}' \label{eq:a1a} \\
&= \int k(\mathbf{x}, \mathbf{x}') k(\mathbf{x}', \mathbf{x}) \, \mathrm{d}\mathbf{x}' \nonumber \\
&\qquad + k(\mathbf{x}, \mathbf{X}) \mathbf{K}^{-1} \left[ \int k(\mathbf{X}, \mathbf{x}') k(\mathbf{x}', \mathbf{X}) \, \mathrm{d}\mathbf{x}' \right] \mathbf{K}^{-1}k( \mathbf{X}, \mathbf{x})   \nonumber \\
&\qquad - 2 k(\mathbf{x}, \mathbf{X}) \mathbf{K}^{-1} \int k(\mathbf{X}, \mathbf{x}') k(\mathbf{x}', \mathbf{x})  \, \mathrm{d}\mathbf{x}'.
\label{eq:a1b}%
\end{align}%
\end{subequations}
If we introduce
\begin{equation}
\hat{k}(\mathbf{x}_1, \mathbf{x}_2) = \int k(\mathbf{x}_1, \mathbf{x}') k( \mathbf{x}', \mathbf{x}_2) \mathrm{d}\mathbf{x},
\label{eq:a2}
\end{equation}
then (\ref{eq:a1b}) can be rewritten as
\begin{align}
\sigma^2(\mathbf{x})\, a_\textit{IVR}(\mathbf{x}) &= \hat{k}(\mathbf{x}, \mathbf{x}) +  k(\mathbf{x}, \mathbf{X}) \mathbf{K}^{-1} \left[ \hat{k}(\mathbf{X}, \mathbf{X}) \mathbf{K}^{-1}k( \mathbf{X}, \mathbf{x}) - 2 \hat{k}(\mathbf{X},  \mathbf{x}) \right]\!.
\label{eq:a3}%
\end{align}%
This shows that to compute IVR and its gradients, we only need a mechanism to compute (\ref{eq:a2}) and its gradients, regardless of the choice of GP kernel.  

For the RBF kernel
\begin{equation}
k(\mathbf{x},\mathbf{x}'; \mathbf{\Theta}) = \sigma_f^2 \exp \!\left[ -(\mathbf{x} - \mathbf{x}')^\mathsf{T} \mathbf{\Theta}^{-1}(\mathbf{x} - \mathbf{x}') /2\right],
\label{eq:s14}
\end{equation}
we have
\begin{subequations}
\begin{equation}
\hat{k}(\mathbf{x}_1, \mathbf{x}_2) = \sigma_f^2 \pi^{d/2} |\mathbf{\Theta}|^{1/2} k(\mathbf{x}_1, \mathbf{x}_2; 2 \mathbf{\Theta})
\end{equation}
and
\begin{equation}
\frac{\mathrm{d}}{\mathrm{d}\mathbf{x}_1}\hat{k}(\mathbf{x}_1, \mathbf{x}_2) = -\hat{k}(\mathbf{x}_1, \mathbf{x}_2) (\mathbf{x}_1 - \mathbf{x}_2)^\mathsf{T} (2\mathbf{\Theta})^{-1}.
\end{equation}
\end{subequations}
For further details, we refer the reader to \cite{mchutchon2013differentiating}.

\section{Analytical Expressions for IVR-LW with RBF Kernel}
\label{app:3}

With the likelihood ratio being approximated with a GMM, the IVR-LW acquisition function becomes
\begin{equation}
a_\textit{IVR-LW}(\mathbf{x}) \approx \frac{1}{\sigma^2(\mathbf{x})} \sum_{i=1}^{n_\mathit{GMM}}\beta_i \, a_i(\mathbf{x}),
\end{equation}
where each $a_i$ is given by
\begin{equation}
a_i(\mathbf{x}) = \int \mathrm{cov}^2(\mathbf{x}, \mathbf{x}') \,\mathcal{N}(\mathbf{x}';\boldsymbol{\omega}_i, \mathbf{\Sigma}_i)  \, \mathrm{d}\mathbf{x}'.
\end{equation}
Using the formula for the posterior covariance, we get
\begin{subequations}
\begin{align}
a_i(\mathbf{x}) &= \int \left[k(\mathbf{x},\mathbf{x}') -k(\mathbf{x},\mathbf{X}) \mathbf{K}^{-1} k(\mathbf{X},\mathbf{x}')\right]^2 \mathcal{N}(\mathbf{x}';\boldsymbol{\omega}_i, \mathbf{\Sigma}_i)  \, \mathrm{d}\mathbf{x}' \label{eq:s214a} \\
&= \int k(\mathbf{x}, \mathbf{x}') k(\mathbf{x}', \mathbf{x})\, \mathcal{N}(\mathbf{x}';\boldsymbol{\omega}_i, \mathbf{\Sigma}_i)  \, \mathrm{d}\mathbf{x}' \nonumber \\
&\qquad + k(\mathbf{x}, \mathbf{X}) \mathbf{K}^{-1} \left[ \int k(\mathbf{X}, \mathbf{x}') k(\mathbf{x}', \mathbf{X})\, \mathcal{N}(\mathbf{x}';\boldsymbol{\omega}_i, \mathbf{\Sigma}_i)  \, \mathrm{d}\mathbf{x}' \right] \mathbf{K}^{-1}k( \mathbf{X}, \mathbf{x})   \nonumber \\
&\qquad - 2 k(\mathbf{x}, \mathbf{X}) \mathbf{K}^{-1} \int k(\mathbf{X}, \mathbf{x}') k(\mathbf{x}', \mathbf{x}) \, \mathcal{N}(\mathbf{x}';\boldsymbol{\omega}_i, \mathbf{\Sigma}_i)  \, \mathrm{d}\mathbf{x}', \\
&= \hat{k}_i(\mathbf{x}, \mathbf{x})  +  k(\mathbf{x}, \mathbf{X}) \mathbf{K}^{-1} \left[  \hat{k}_i(\mathbf{X}, \mathbf{X}) \mathbf{K}^{-1}k( \mathbf{X}, \mathbf{x}) - 2  \hat{k}_i(\mathbf{X}, \mathbf{x}) \right]\!,
\label{eq:s214b}%
\end{align}%
\end{subequations}
where we have defined 
\begin{equation}
\hat{k}_i(\mathbf{x}_1, \mathbf{x}_2) = \int k(\mathbf{x}_1, \mathbf{x}') k( \mathbf{x}', \mathbf{x}_2) \, \mathcal{N}(\mathbf{x}';\boldsymbol{\omega}_i, \mathbf{\Sigma}_i)  \, \mathrm{d}\mathbf{x}'.
\label{eq:s213}
\end{equation}
Therefore, to evaluate $a_i$ and its gradients, we only need a mechanism to compute $\hat{k}_i$ and its gradients.  

For the RBF kernel \eqref{eq:s14}, it is straightforward to show that 
\begin{subequations}
\begin{equation}
\hat{k}_i(\mathbf{x}_1, \mathbf{x}_2) = |2 \mathbf{\Sigma}_i \mathbf{\Theta}^{-1} + \mathbf{I}|^{-1/2} k(\mathbf{x}_1, \mathbf{x}_2; 2 \mathbf{\Theta}) k(\mathbf{x}_1 + \mathbf{x}_2, \boldsymbol{\omega}_i; \mathbf{\Theta} + 2 \mathbf{\Sigma}_i)
\label{eq:s215}
\end{equation}
and
\begin{equation}
\frac{\mathrm{d}\hat{k}_i(\mathbf{x}_1, \mathbf{x}_2)}{\mathrm{d}\mathbf{x}_1}  = \hat{k}_i(\mathbf{x}_1, \mathbf{x}_2) \left\{ -\mathbf{x}_1^\mathsf{T} \mathbf{\Theta}^{-1} + \frac{1}{2}\left[\boldsymbol{\omega}_i^\mathsf{T} + (\mathbf{x}_1+\mathbf{x}_2)^\mathsf{T}\mathbf{\Theta}^{-1} \mathbf{\Sigma}_i \right] (  \mathbf{\Sigma}_i + \mathbf{\Theta} / 2 )^{-1}\right\}\!.
\label{eq:s216}
\end{equation}
\end{subequations}
For further details, we refer the reader to \cite{mchutchon2013differentiating}.

\end{document}